\documentclass[10pt,twocolumn,letterpaper]{article}

\usepackage{xcolor} 

 \usepackage{iccv}              
\usepackage{colortbl}
\usepackage[accsupp]{axessibility}  
\usepackage[linesnumbered, ruled]{algorithm2e}
\definecolor{iccvblue}{rgb}{0.21,0.49,0.74}
\usepackage[pagebackref,breaklinks,colorlinks,allcolors=iccvblue]{hyperref}

\def\paperID{1621} 
\def\confName{ICCV}
\def\confYear{2025}

\title{Penalizing Boundary Activation for Object Completeness in \\ Diffusion Models}
\author{
Haoyang Xu\textsuperscript{1} \and
Tianhao Zhao\textsuperscript{1} \and
Sibei Yang\textsuperscript{2} \and
Yutian Lin\textsuperscript{1}\thanks{Corresponding author.}\and \\
\textsuperscript{1}School of Computer Science, Wuhan University \\
\textsuperscript{2}School of Computer Science and Engineering, Sun Yat-sen University \\
{\tt\small \{haoyxu, happytianhao, yutian.lin\}@whu.edu.cn} \\
{\tt\small yangsb3@mail.sysu.edu.cn}
}

\begin{document}
\maketitle
\begin{abstract}
Diffusion models have emerged as a powerful technique for text-to-image (T2I) generation, creating high-quality, diverse images across various domains. However, a common limitation in these models is the incomplete display of objects, where fragments or missing parts undermine the model's performance in downstream applications.
In this study, we conduct an in-depth analysis of the incompleteness issue and reveal that the primary factor behind incomplete object generation is the usage of \textit{RandomCrop} during model training. This widely used data augmentation method, though enhances model generalization ability, disrupts object continuity during training. To address this, we propose a training-free solution that penalizes activation values at image boundaries during the early denoising steps. Our method is easily applicable to pre-trained Stable Diffusion models with minimal modifications and negligible computational overhead. Extensive experiments demonstrate the effectiveness of our method, showing substantial improvements in object integrity and image quality. Code is available at \url{https://github.com/HaoyXu7/Object_Completeness}. 
\end{abstract}    
\section{Introduction}
\label{sec:intro}
\begin{figure}[t]
    \centering
    \footnotesize
    \tabcolsep=.1mm
    \begin{tabular}{p{14.5mm}p{17mm}p{17mm}p{17mm}p{17mm}}
        \raisebox{2mm}{\parbox{14.5mm}{\centering\textbf{\quad}}}&
        \raisebox{2mm}{\parbox{16mm}{\centering\textit{A bed}}}&
        \raisebox{2mm}{\parbox{16mm}{\centering\textit{A red car}}}&
        \raisebox{2mm}{\parbox{16mm}{\centering\textit{A \textbf{\underline{complete}} bathtub}}}&
        \raisebox{2mm}{\parbox{16mm}{\centering\textit{A cap \textbf{\underline{in the middle }}}}}
        \\
        \raisebox{7mm}{\parbox{14.5mm}{\centering\textbf{SDv2.1}}}
        &
        \includegraphics[width=16mm]{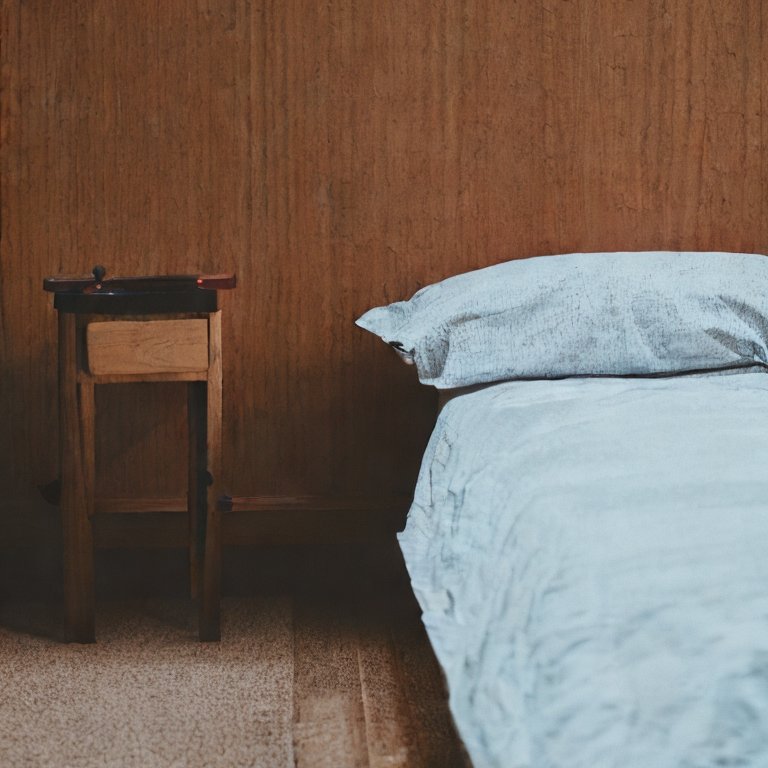}           &
        \includegraphics[width=16mm]{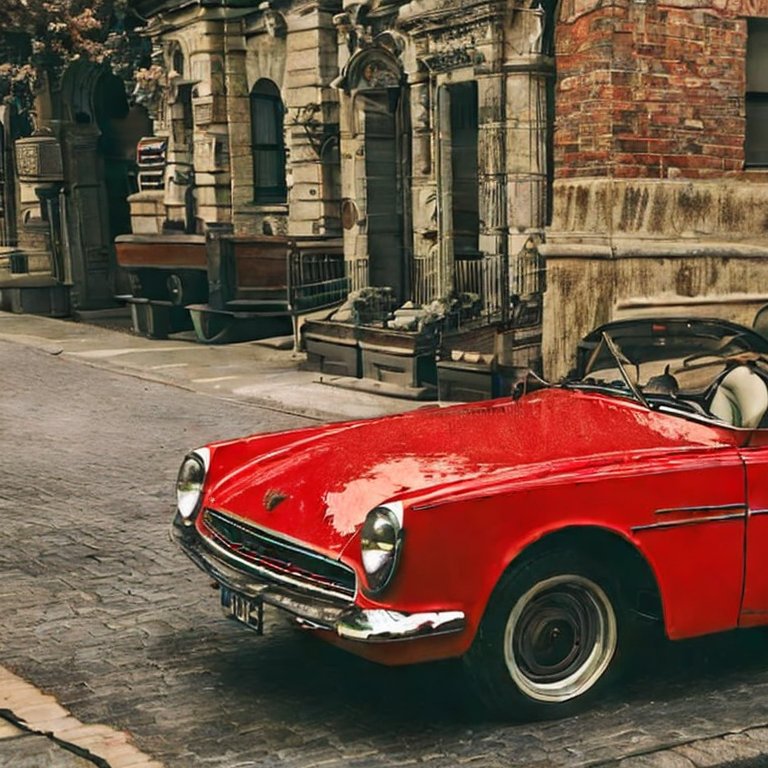} &
        \includegraphics[width=16mm]{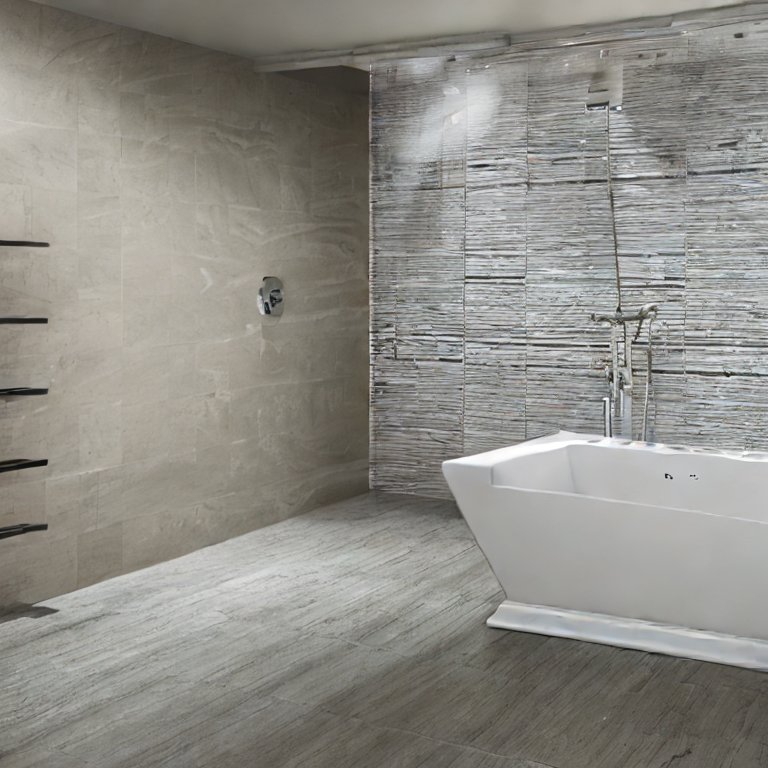} &
        \includegraphics[width=16mm]{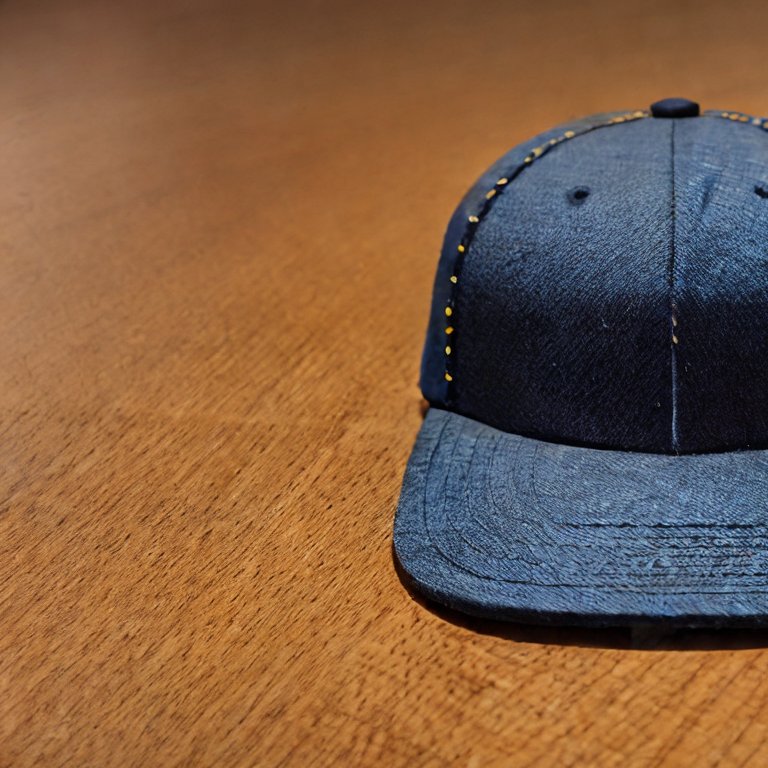} \\
        \raisebox{7mm}{\parbox{14.5mm}{\centering\textbf{SDv1.5}}}
        &
        \includegraphics[width=16mm]{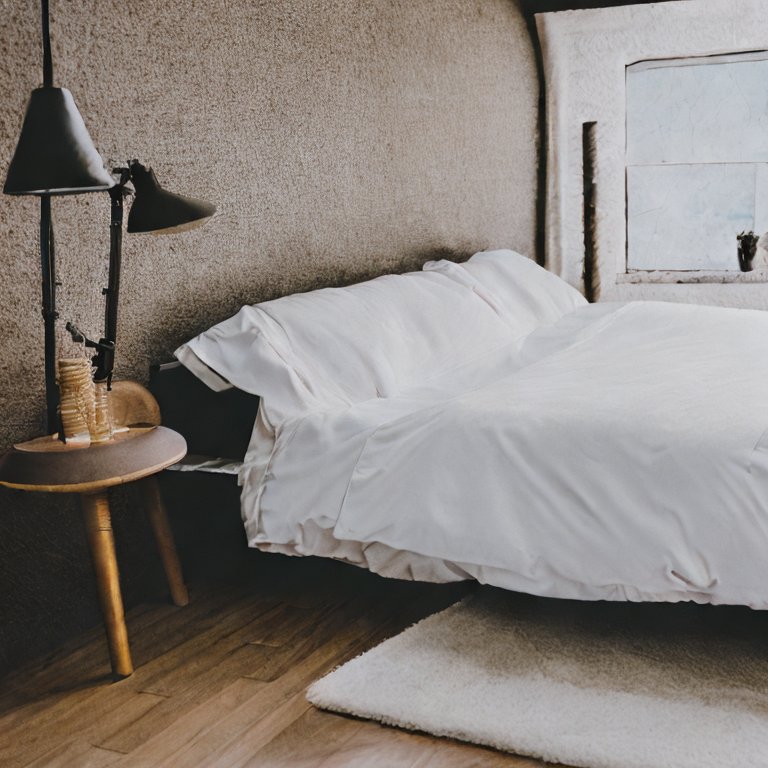}       &
        \includegraphics[width=16mm]{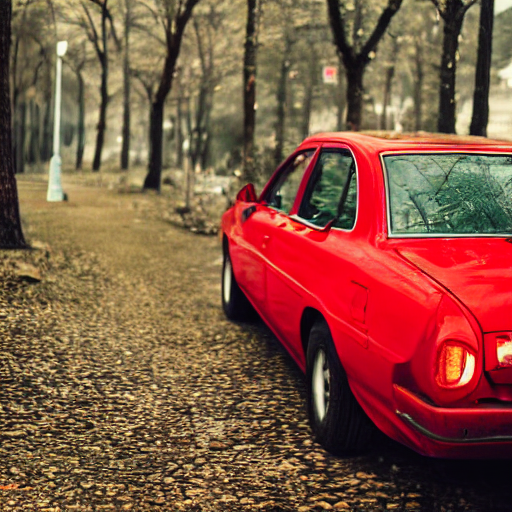}                         &
        \includegraphics[width=16mm]{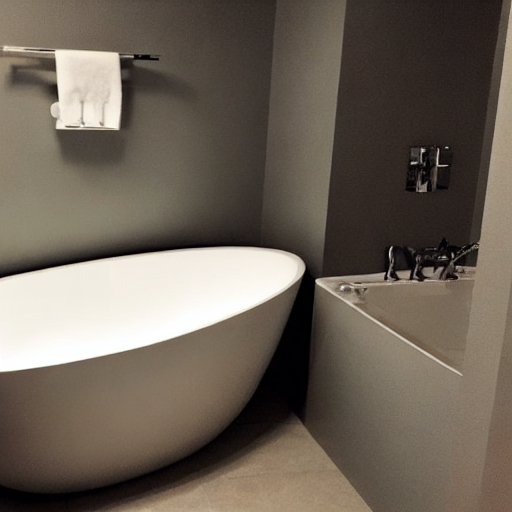} &
        \includegraphics[width=16mm]{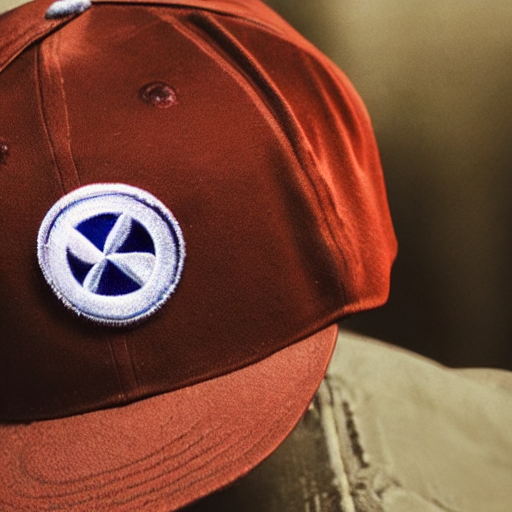}                                                                                                                                                  \\
    \end{tabular}
    \vspace{-1mm}
    \caption{
    Incomplete objects generated by different Stable Diffusion models with varying prompts, even with modifiers indicating completeness ($\eg$ ``complete'', ``in the middle'').
    }
    \label{fig:result1}
\end{figure}

Diffusion models excel in text-to-image (T2I) generation by producing high-quality, diverse outputs~\cite{ho2020denoising, song2020score, ramesh2021zero,wang2024silent, rombach2022high, saharia2022photorealistic}, with broad applications in areas like dataset synthesis and video generation~\cite{meng2021sdedit, dhariwal2021diffusion, whang2022deblurring, lugmayr2022repaint, kawar2022jpeg, ruiz2023dreambooth}. 
However, the generated images often contain incomplete targets – fragmented or missing object parts as shown in ~\cref{fig:result1}. 
Such object incompleteness issues not only undermine the reliability of diffusion models but also adversely affect the performance of downstream tasks dependent on the generated data.



However, this issue has largely been overlooked in existing studies~\cite{ren2024move,shi2024dragdiffusion,liu2024drag,mou2023dragondiffusion}. Some studies~\cite{rassin2024linguistic,sueyoshi2024predicated,hertz2022prompt} dismiss these minor imperfections as inherent randomness in the generation process, while others~\cite{chen2024training,xie2023boxdiff,li2023gligen,mao2023guided} categorize them as generation failures without delving into the underlying causes.
The layout generation methods~\cite{li2023gligen,zheng2023layoutdiffusion,chen2024training} benefit from planning the placement of objects to address the object incompleteness issue. 
These methods either produce objects that overflow the layout due to insufficient constraints or apply overly strict constraints, causing distortion. 
SDXL~\cite{podell2023sdxl} recognizes this issue and attempts to mitigate it by incorporating the coordinates of the cropped operation during training. While the method does not eliminate the issue and carries the risk of introducing errors, 
such as the unnatural clustering of multiple objects at the center of the image.
Despite these efforts, there has been little in-depth exploration of the underlying causes of the incompleteness problem or further advances in solutions. 

To better understand the issue, we conducted an in-depth investigation into the phenomenon of generative incompleteness and its causes. 
We manually check that the incompleteness issue persists universally across various diffusion models, as shown in~\cref{tab:fact2}, with the highest object incompleteness rate (from DALL-E 2) reaching 47.3\% . Even SDXL, which is specifically trained to address this issue, still exhibits an incompleteness rate of 18\%. We attempt to adjust prompts by adding adjectives to specify object completeness. However, as shown in the left two columns of ~\cref{tab:fact}, this does not resolve the issue, highlighting that \textit{the issue lies beyond prompt engineering}.
Based on the findings above, we propose two primary hypotheses: the issue might stem from inherent incompleteness in the object characteristics present in the training dataset or from biases introduced by the training strategy. 
We design experiments to establish that 
the usage of \textit{RandomCrop} as a data augmentation method is the primary factor contributing to this issue. 
More details of the discussions of the two hypotheses are illustrated in Sec.~\ref{sec:ca}.
\begin{table}[t]
    \centering
    \footnotesize
    \resizebox{0.98\linewidth}{!}{
    \begin{tabular}{lccccc}
        \toprule
        \textbf{Model}  & DALL-E 2~\cite{Dayma_DALL·E_Mini_2021} &  SDv2~\cite{rombach2022high}   &  SDv3~\cite{rombach2022high} & SDXL~\cite{podell2023sdxl} \\
        \midrule 
        \textbf{HOIR}   &  47.3\%    & 45.5\%   &30.1\%  &18.0\%\\
        \bottomrule
    \end{tabular}
    }
    \caption{Statistics on the incompleteness of objects of different T2I models using datasets constructed from benchmarks in ~\cref{tab:fact}. \textbf{HOIR} is the abbreviation for human-evaluated object incompleteness rate.}
    \label{tab:fact2}
    \vspace{-8mm}
\end{table}

Despite its drawbacks, \textit{RandomCrop} remains an indispensable component of diffusion model training, as it significantly enhances the model's diversity and generalization capabilities. 
Moreover, addressing the incompleteness issue by fine-tuning diffusion models without \textit{RandomCrop} would be prohibitively expensive due to the high costs involved in training. 
Thus we propose a training-free solution to achieve object completeness in generic generation scenarios. To achieve this, we implicitly guide the subject toward the central region by reducing the probability of generating objects near the image edges. Specifically, we first utilize cross-attention maps to identify corresponding self-attention maps. These features are then integrated to penalize activation values at image boundaries during the early stages of the denoising process. 
This way, we can facilitate the generation of complete objects during model inference without requiring additional model training or external spatial signals, offering a lightweight and efficient solution.
Our approach is inherently orthogonal to SDXL's training methodology while resolving its potential layout challenges and can be easily applied to existing diffusion models with minimal modifications and computational overhead. 



In summary, our contributions are threefold: 
\begin{itemize}
    \item We identify and validate the cause of incomplete object generation in diffusion models, attributing it primarily to the use of \textit{RandomCrop} during model training. 
    
    \item We propose a training-free method that improves object completeness in diffusion-generated images by reducing the probability of generating objects near the image edges, easily applicable to pretrained diffusion models with minimal additional cost.
    
    \item Experimental results validate the effectiveness of our method. Our method reduces the object incompleteness rate of SDv2.1 to 17.3\% with a negligible computation cost, demonstrating substantial improvements in image quality and object completeness, making diffusion models more reliable for applications.
\end{itemize}

\begin{table}[t]
    \centering
    \footnotesize
    \resizebox{0.98\linewidth}{!}{
    \begin{tabular}{lc|lc}
        \toprule
        \textbf{Prompt Type}  & \textbf{HOIR} & \textbf{Prompt Source}  & \textbf{HOIR}\\
        \midrule 
        a [CLASS]   &  45.7\%    & COCO Caption   & 38.5\%  \\
        a complete [CLASS]   &  43.2\%  & Flickr30k   & 40.3\%  \\
        a [CLASS] in the middle   &  44.3\%     & DrawBench   & 22.5\% \\
        \bottomrule
    \end{tabular}
    }
    \caption{Statistics on the incompleteness rate of the images generated by SDv2.1 across different prompts. We evaluate both manually constructed prompts and public datasets. \textbf{HOIR} is the human-evaluated abbreviation for object incompleteness rate.}
    \label{tab:fact}
    \vspace{-2mm}
\end{table}
\section{Related Work}
\label{sec:formatting}

\subsection{Controlled Generation in Diffusion}
Many studies~\cite{nichol2021glide, gu2022vector, ramesh2022hierarchical, harvey2022flexible, yang2023diffusion,wang2023diffusion} aim to enable diffusion models to generate high-quality results under more precise control. Among these, controlled generation~\cite{xu2018attngan, huang2024robin,yu2018generative, parmar2023zero, wang2024compositional,huang2025implicit} is frequently used to modify the attributes and details of images, while maintaining the model's generative capabilities. These methods leverage the attention mechanism in the generation process~\cite{ronneberger2015u}. 
Some studies~\cite{chefer2023attend,hertz2022prompt} incrementally updates the activation levels of specific words corresponding to objects in $A$ to control the expression intensity of the prompt. Others ~\cite{rassin2024linguistic,sueyoshi2024predicated} uses cross-attention maps to bind entities and attributes, thereby ensuring the generated results more accurately align with the prompt. 

Additionally, with the release of pre-trained models like Stable Diffusion~\cite{rombach2022high}, many tasks find it more appropriate to transfer the prior knowledge of these models rather than training the network~\cite{ruder2019transfer, brown2020language, nichol2021improved}. FastComposer~\cite{xiao2024fastcomposer} introduces cross-attention maps for the localization of multiple specific subjects in the generation, ensuring that the attention on reference subjects is precisely focused on the corresponding regions of the target image.
MasaCtrl~\cite{cao2023masactrl} obtains the object mask from the cross-attention map and applies it in the proposed mutual self-attention strategy to control the generation results.

\subsection{Seed Select in Latent Space} 
The Gaussian noise as the starting point of the diffusion denoising process may differ. As a result, even with the same prompt, the generated outcome can vary~\cite{ho2020denoising, song2019generative, salimans2022progressive}. SeedSelect~\cite{samuel2024generating} solves the problem of difficulty in generating specific objects by selecting suitable seeds in the noise space. ~\citet{mao2023guided} discovered that certain pixel blocks in the initial latent image have a tendency to generate specific content. They modify these blocks to gain precise control over specific regions of the image. S2ST~\cite{greenberg2023s2st} inverses the original image into a noise seed and matches the target scene through seed translation and trajectory optimization to control the style and layout of the result. Consequently, they achieved a relatively controllable generation process by attempting to select seeds in the seed space that better aligned with their objectives.

\subsection{Layout Control in Image Generation}
The prompt exhibits ambiguity in guiding the image layout. Layout generation attempts to specify the generation regions of objects through additional positional information (such as box, keypoints, depth map, semantic map, etc.). GLIGEN~\cite{li2023gligen} incorporates new conditional information into the generation process through a gated self-attention layer. LayoutDiffusion~\cite{zheng2023layoutdiffusion} integrates the bounding box conditions of multiple objects into the proposed layout fusion module and then combines it with cross-attention to precisely control spatial information. ~\citet{chen2024training} use cross-attention maps to update the current step latent and participate in the next iteration. Some studies~\cite{balaji2022ediff,xie2023boxdiff} directly increase the probability of specific objects in the attention map with masks to achieve layout arrangement. 
Although layout generation attempts to address the completeness of object generation from a different angle, it sometimes places objects in physically unrealistic poses and still fails to address object incompleteness. Our method avoids such placement errors by adhering to the diffusion setup and effectively addresses incompleteness.


\section{Object Incompleteness}
\label{sec:oi}

This section analyzes persistent object incompleteness in diffusion models (\cref{sec:is}). We further investigate the causes, focusing on the dataset characteristics and data augmentation techniques (\cref{sec:ca}).
\subsection{Issue Statement}
\label{sec:is}
To investigate the issue, we test prompts containing a single object (such as a backpack, a car, \etc), along with various T2I model benchmarks, obtaining the corresponding image. \cref{fig:result1} illustrates some cases, where the generated images only display a part of the object ($\eg$ the rear half of the car is missing, the bathtub and cap are missing their right halves). The situation remains the same when we use prompts with spatial descriptions. 

To further investigate the occurrence of this issue, we randomly select 1,200 object categories (including animals, furniture, tools, clothing, vehicles, \etc), and generate corresponding prompts for each category. At the same time, we also prompt SDv2.1 with a randomly sampled subset of 1,200 images from each of COCO Caption~\cite{chen2015microsoft} and Flickr30k~\cite{young2014image}, as well as the full DrawBench~\cite{saharia2022photorealistic}. For all results, we manually evaluate an image complete if its subject is mostly intact. 
As shown in ~\cref{tab:fact}, the probability of the object incompleteness issue occurring is surprisingly 45.7\% when the given prompt type is ``a [CLASS]". Even after adding adjectives indicating completeness to the prompt, the probability only slightly decreased to 43\%. 
Using DrawBench as the prompt source yields the lowest HOIR of 22.5\%, but with only 200 prompts, the results may be unstable.

\begin{figure}[t]
    \includegraphics[width=\linewidth]{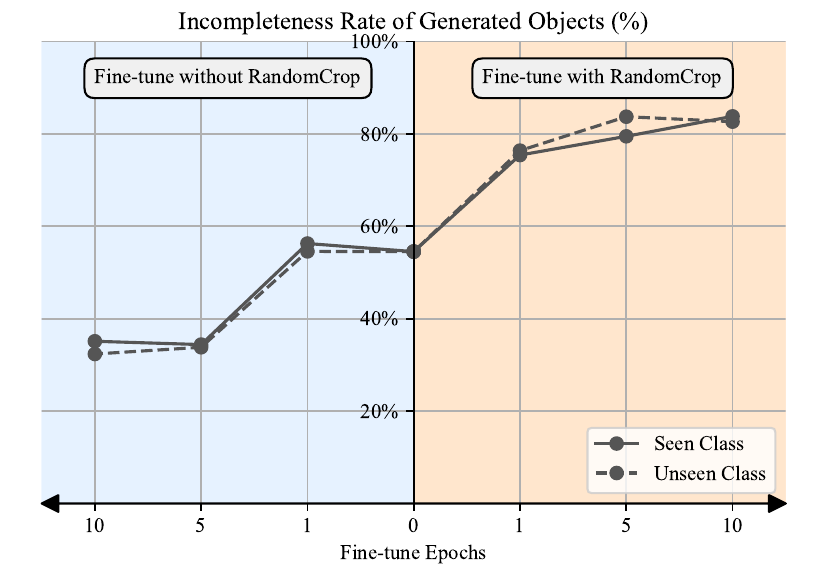} 
    \vspace{-2mm}
    \caption{The trend of the rate of object incompleteness in images generated by diffusion models fine-tuned in different ways.
    Here, ``Seen Class'' and ``Unseen Class'' indicate whether the category of the object generated by the fine-tuned diffusion model is seen in the fine-tuning data.
    }
    \label{fig:augmentation_analysis}
    \vspace{-2mm}
\end{figure}

\subsection{Cause Analysis}
\label{sec:ca}
Regarding the underlying causes, we posit that images generated by diffusion models exhibit statistical distribution alignment with the input tensors during training.  
Therefore, the issue of object incompleteness likely exists in the training input.
This issue may arise from two main factors: 1) the dataset, and 2) data augmentation.
Below we discuss the potential causes of this issue from these two perspectives.\\
\textbf{The dataset.}
We select a collection of datasets employed in training diffusion models, including the aesthetic subset of LAION-5B~\cite{schuhmann2022laion}, and then we manually calculate the incompleteness rate of 1,200 randomly sampled images from these datasets to estimate the overall incompleteness rate of the entire dataset. However, we find it to be only 4\%, much lower than the probability observed in the images generated by the diffusion model. This suggests that the primary cause of the issue does not stem from the dataset itself.\\
\textbf{Data augmentation.}
The images in the dataset undergo data augmentation to increase diversity before being input into the diffusion model in the training process. We find that the data augmentation technique \textit{RandomCrop}, which is widely used in various diffusion models for training, may contribute to incomplete objects. To validate the impact of \textit{RandomCrop} on this issue, we separately fine-tune SDv2.1 on 10,000 images selected from the aesthetic subset of the LAION-5B dataset, using the original images and their \textit{RandomCrop}-augmented versions.

As shown in ~\cref{fig:augmentation_analysis}, \textit{RandomCrop} significantly increases the probability of object incompleteness in images generated by the fine-tuned model. Moreover, as the training epochs increase, the incompleteness rate exhibits a monotonically increasing trend and stabilizes around epoch 5. In contrast, the incompleteness rate of the fine-tuned model with original images monotonically decreases with more fine-tuning epochs. To demonstrate that this tendency toward incompleteness or completeness is not caused by the content of the images that we select, we divide the objects in the generated images into two categories based on whether they appeared in the fine-tuning dataset and calculate the probability of incompleteness for each category. As shown in the ~\cref{fig:augmentation_analysis}, the trends and values of the dashed line (representing unseen categories) and the solid line (representing seen categories) are largely consistent. This indicates \textit{RandomCrop} is a primary cause of the incompleteness issue. 

\section{Penalizing Boundary Activation}

According to the analysis in \cref{sec:oi}, the issue of object incompleteness in images generated by diffusion models stems from the use of \textit{RandomCrop}. However, \textit{RandomCrop} plays a crucial role in improving the model’s generalization ability and enhancing the diversity of its generated images, making it indispensable. Furthermore, existing pre-trained diffusion models, such as Stable Diffusion 2.1, require substantial computational resources, making it impractical to fine-tune solely to address this issue.

To address the incompleteness issue, we propose a training-free method that implicitly guides the subject toward the central region by reducing the probability of generating objects near the image edges. And our method can be applied to all existing diffusion models with minimal modifications and negligible additional computational cost. 

\begin{figure}[t]
    \centering
    \includegraphics[width=\linewidth]{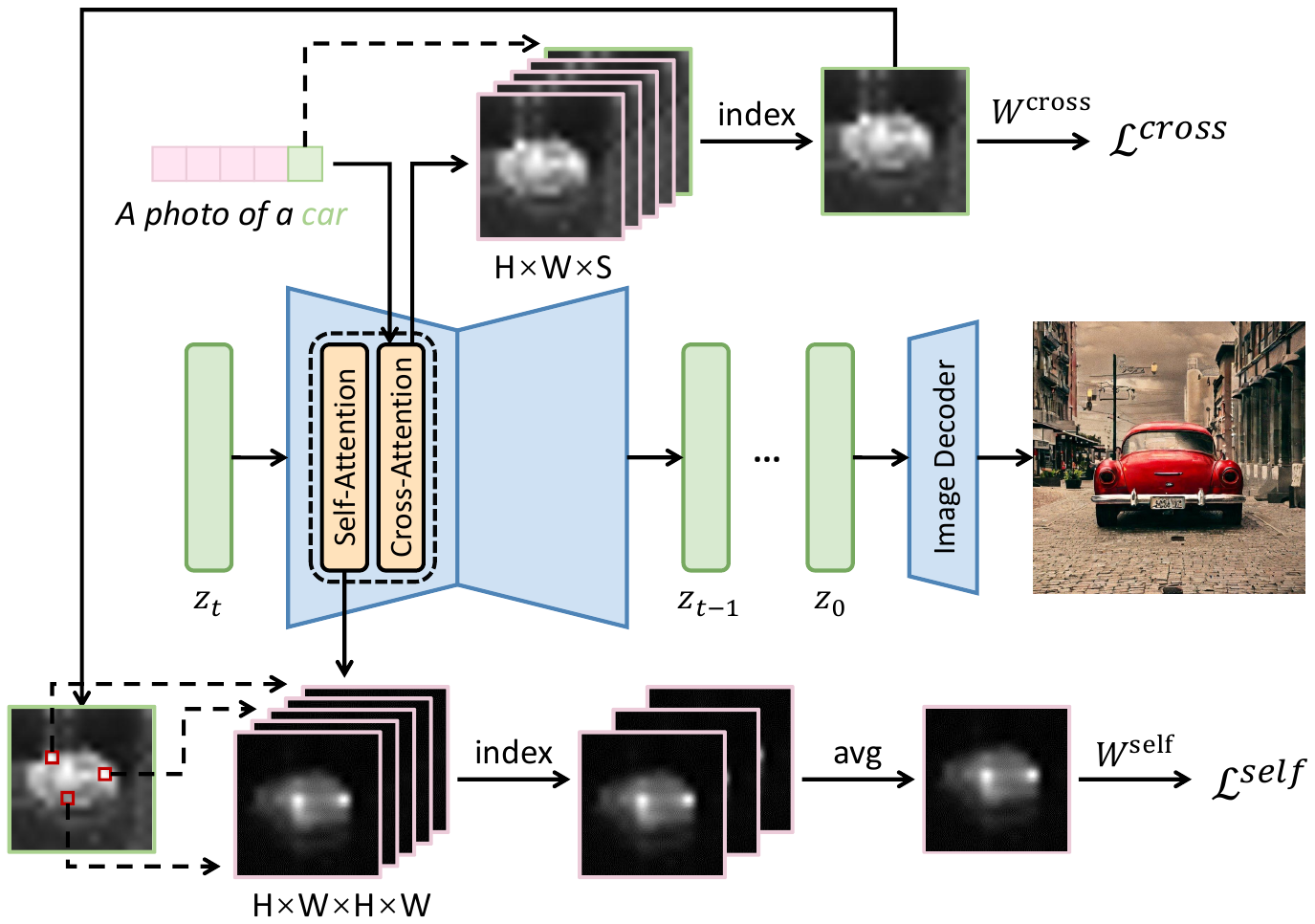}
    \vspace{-1mm}
    \caption{An overview of our method with cross-attention constraint and self-attention constraint. The cross-attention maps, which encode semantic information, are initially extracted and subsequently processed using a clustering algorithm to identify the corresponding self-attention maps that represent spatial information. These two are then integrated via the dispelling constraint during the early denoising stage, effectively driving objects away from boundaries.}
    \vspace{-3mm}
    \label{fig:method_overview}
\end{figure}

The overview of the method is shown in Fig.~\ref{fig:method_overview}.
Our method leverages cross-attention maps to identify corresponding self-attention maps. These attention maps are integrated to construct the dispelling constraint to optimize the diffusion latent during the early denoising stage, effectively driving objects away from boundaries. ~\cref{sec:constraint} introduces our proposed unified dispelling constraint and ~\cref{sec:Optimization} details the optimization strategy.

\subsection{Dispelling Constraint}
\label{sec:constraint}
Our method aims to reduce the probability of generating objects near the image edges. Therefore, we utilize cross-attention maps to suppress semantic expression at the boundaries while leveraging self-attention to adjust spatial layout. According to our findings, when we focus on only one of these two attention mechanisms, the image modification progresses insufficiently in the desired direction and requires more iterations to optimize, leading to a decline in image quality. When constrained solely by $M^{\texttt{cross}}$, the results still struggle to capture the spatial structure accurately and remain incomplete. Conversely, when only $M^{\texttt{self}}$ is employed, the generated images tend to lack fine-grained details and exhibit deficiencies in semantic consistency. So we combine both attention maps to drive objects away from the boundaries. And our approach facilitates efficient control of object completeness during the critical initial steps, which are pivotal for shaping the coarse structure~\cite{wang2023diffusion}.\\
\textbf{Extracting Cross-Attention Maps.} Cross-attention maps are used to model the interactions between the prompt and the image. So we first extract the cross-attention maps $M^{\texttt{cross}}$ corresponding to the objects $X$ in the prompt, which represents the semantic imprint of $X$ within the image. In other words, it reflects the intensity of the location in the $M^{\texttt{cross}}$ associated with the object $X$. \\
\textbf{Searching Self-Attention Maps.} Although the attention $M^{\texttt{cross}}$ is effective in capturing the object contours, it focuses more on the semantic information of the image rather than spatial information. Therefore, to better ensure the completeness of objects in the image, we introduce the self-attention maps $M^{\texttt{self}}$ that encode spatial reference information of the image. When capturing self-attention maps, we apply Gaussian smoothing to $M^{\texttt{cross}}$ to avoid noise caused by extremum Point, making the variation of local values in object $X$ relatively uniform. Subsequently, we use a clustering algorithm to identify \( K \)  key points to guide self-attention, ensuring that the selected points are well distributed. The \( K \)  key points are selected as the top \( K \) (indexed by $p_1,p_2,\dots,p_K$) cluster centers after clustering. Afterward, we find the corresponding self-attention maps $M^{\texttt{self}}$ for these points $M^{\texttt{self}}_{p_k}$ and compute their average to obtain the final self-attention map.\\
\textbf{Dispelling Loss.}
The core of our constraint is to implicitly guide the subject toward the central region by reducing the probability of generating objects near the image edges. Specifically, we design a loss function on the attention maps, where the activation values of a randomly selected internal region $A_{inter}$ are subtracted from those of the surrounding region $A_{sur}$.  $\mathcal L $ is computed as follows:
\begin{equation} 
\mathcal L = \alpha \cdot A_{sur} - \beta \cdot A_{inter} ,
\label{eq:important}
\end{equation}
where $\alpha$ and $\beta$ are hyperparameters.
In summary, our method integrates both types of guidance. A detailed visual comparison and analysis are provided in the ~\cref{sec:Ablation}.

\begin{algorithm}[t]

\caption{Pseudocode of our method}
\label{alg:Dual-attention}
\KwIn{An object category $X$ and corresponding text prompt $\mathcal{P}$,  a timestep $t$, time thresholds $T_1$, hyperparameter $\alpha_t$, $K$, and a pre-trained Stable Diffusion model $SD$.}
\KwOut{Latent representation $z_{t-1}$ for the next timestep}

$ \_, M^{\texttt{cross}}, M^{\texttt{self}} \gets SD(z_t, \mathcal{P}, t)$\;
$x\gets  \text{get\_index}(X,\mathcal{P})$\;
$\mathcal L^{\texttt{cross}} \gets M^{\texttt{cross}}_x$\;
$\ (p_1,p_2,\dots,p_K)\gets \text{topk}(M^{\texttt{cross}}_x,K)$\;
$M^{\texttt{self}}_{\texttt{avg}}\gets \frac{1}{K}\sum_{k=1}^{K} M^{\texttt{self}}_{p_k}$\;
$\mathcal L^{\texttt{self}} \gets M^{\texttt{self}}_{\texttt{avg}}$\;
\If{$ t > T_1 $}{
    $\mathcal L^{\texttt{all}}\gets \mathcal L^{\texttt{cross}}+\mathcal L^{\texttt{self}}$\;
}
$z_t' \gets z_t - \alpha_t \cdot \nabla_{z_t} \mathcal{L^{\texttt{all}}}$\;
$z_{t-1}, \_, \_ \gets SD(z_t', \mathcal{P}, t)$\;
\Return{$z_{t-1}$}\; 

\end{algorithm}

\begin{table*}[t]
    
    \footnotesize
    \centering
    \tabcolsep=1.5mm
    \resizebox{\linewidth}{!}{
    \begin{tabular}{lc|cc|c|c|ccc}
        \toprule
        \textbf{Method} & \textbf{LLM-based}  & {\textbf{HOIR $\downarrow$} } & \textbf{LOIR $\downarrow$} & \textbf{Time Cost (s) $\downarrow$} & \textbf{CLIP-IQA $\uparrow$} 
        & \textbf{PickScore $\uparrow$} & \textbf{Hpsv2 $\uparrow$} & \textbf{ImageReward $\uparrow$}
         \\
        \midrule
        Stable Diffusion~\cite{rombach2022high} & $\times$   & 45.7 \%  & 32.0  \% & 5.51 & \textbf{0.714}& 20.03& 25.20 & 0.221\\
        \midrule
        GLIGEN~\cite{li2023gligen}      & $\times$      & 34.2 \%     & 27.9  \%   & 8.74  & 0.672& 20.59& 24.73& 0.177\\
        Layout Guidance~\cite{chen2024training}& $\times$  & 42.1 \%       & 35.7  \%      & 24.02 & 0.654& 22.81& 24.18& -0.222\\
        Boxdiff~\cite{xie2023boxdiff}     &  $\times$     & 39.1  \%    & 29.8  \%      & 7.18 & 0.674& 20.21& 23.14 & -0.134\\
        LayoutLLM~\cite{qu2023layoutllm}     & \checkmark        & 32.7  \%    & 28.6  \%   & 10.44  & 0.594& 22.46& 24.98 & 0.109\\
        RPG~\cite{yang2024mastering}        &\checkmark    & 35.4  \%    & 30.5  \%     & 8.54 & 0.653& 22.57& 25.15 & 0.240\\
        LayoutGPT~\cite{feng2023layoutgpt}   &\checkmark     & 30.3  \%    & 31.5  \%   & 14.36 & 0.631& 21.94& 24.83 & 0.175\\
        SLD~\cite{wu2024self}            &\checkmark        & 27.1  \%    & 23.1  \%      & 9.54  & 0.694& 23.07& 25.17 & 0.253\\
        \midrule
        \textbf{Ours}    &   $\times$     &  \textbf{17.3 \%}  & \textbf{11.7 \%}  & \textbf{5.75} & 0.703 & \textbf{23.41} & \textbf{25.66} & \textbf{0.327}\\

        \bottomrule
    \end{tabular}}
    \vspace{-1mm}
    
    \captionof{table}{Quantitative comparison between the layout-based methods and our method across different metrics, where \textbf{H (L)OIR} is the abbreviation for Human (LMM)-Evaluated object incompleteness rate. All methods are conducted using \textbf{Stable Diffusion version 2.1}. In the Time Cost metric, the overhead induced by the Large Language Model (LLM) is excluded.}
    \label{tab:experiments}
    \vspace{-4mm}
\end{table*}

\subsection{Latent Representation Optimization}
\label{sec:Optimization}
We have computed two distinct losses, \(\mathcal L^{\texttt{cross}} \) and \(\mathcal L^{\texttt{self}} \) through ~\cref{eq:important}, based on the dispelling constraints. Since we aim to improve the completeness of objects, we optimize the latent representation \( z_t \) through gradient backpropagation to minimize both losses:
\begin{equation}
    z'_t = z_t - \alpha_t \cdot \nabla_{z_t} (\mathcal L^{\texttt{cross}}+\mathcal L^{\texttt{self}}),
\end{equation}
where $\nabla_{z_t}$ denotes the magnitude of the gradient updates, which gradually decreases over time. 

Specifically,  we set an early timestep $ T_1 $ $(T>T_1>0)$ ($T_1 = 45$ when $T = 50$ ) and apply the unified dispelling constraint for timesteps $t$ \( (T\geq t > T_1) \). All module parameters are fixed, and we adapt the diffusion latent only during inference time.

The pseudocode of our method is shown in \cref{alg:Dual-attention}.

\section{Experiments}

\subsection{Experimental Setup} 
\textbf{Datasets.}\label{sec:data}
For this study, we construct two prompt datasets. One dataset consists of 1,200 prompts from various categories, carefully selected to represent medium to small-sized physical objects. These objects were sourced from the ``lists of lists" on Wikipedia and include categories such as animals, stationery, household items, furniture, vehicles, clothing, and fruits and vegetables. The other dataset consists of 30,000 prompts randomly selected from COCO Caption~\cite{chen2015microsoft}, Flickr30k~\cite{young2014image} and Drawbench~\cite{saharia2022photorealistic}, and excludes prompts containing more than five objects. \\
\textbf{Implementation details.} In our experiment, unless otherwise specified, the reported results are obtained by applying our method to SDv2.1 and SDv1.5. During the forward process, we perform a total of 50 diffusion steps, but our method is applied only to the first 5 steps. For the parameters, we set $\alpha = 1.2$, $\alpha = 0.4$
We extract attention maps using 16 × 16 attention units, as previous work~\cite{hertz2022prompt} has shown that this configuration captures the most semantic information. To select activation values from the cross-attention maps, we set \( K = 10 \), which typically ensures that the object’s spatial information is fully captured. 
Regarding the matrices \( W^{\texttt{cross}} \) and \( W^{\texttt{self}} \), which match the size of the extracted attention maps, we modify specific areas.

\begin{figure*}[t]
    \tabcolsep=0.2mm
    \scriptsize
    \centering
    \resizebox{\linewidth}{!}{
    \begin{tabular}{c@{\hspace*{1mm}}ccc@{\hspace*{2mm}}ccc@{\hspace*{2mm}}ccc}
                                                                                           &
        \multicolumn{3}{c}{\emph{A backpack}}                                    &

        \multicolumn{3}{c}{\emph{A car}}                                        &
        \multicolumn{3}{c}{\emph{A bathtub}}                                       \\
        \rotatebox[origin=l]{90}{\parbox{13mm}{\centering Stable Diffusion}}               &
        \includegraphics[width=14mm]{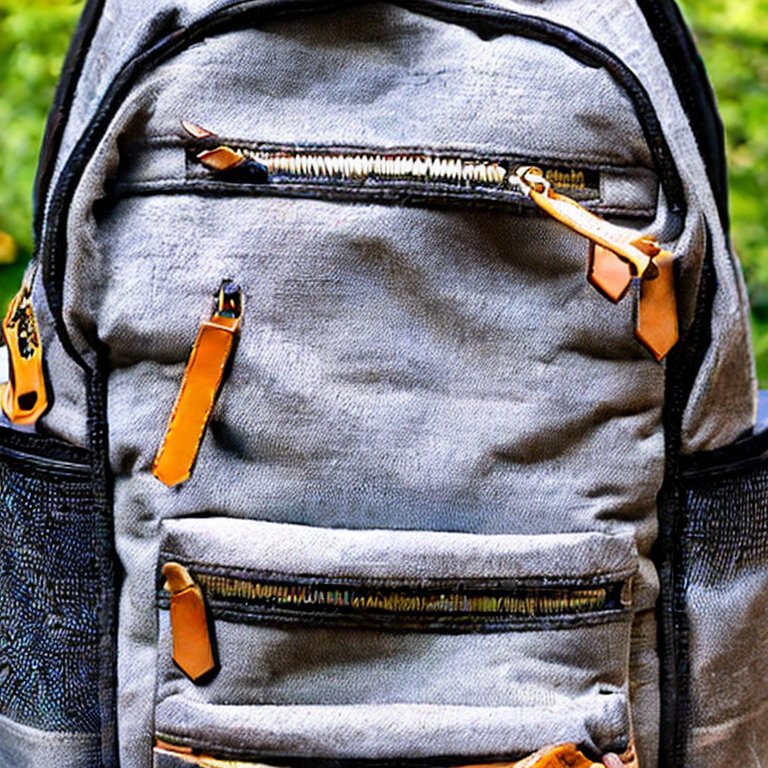}     &
        \includegraphics[width=14mm]{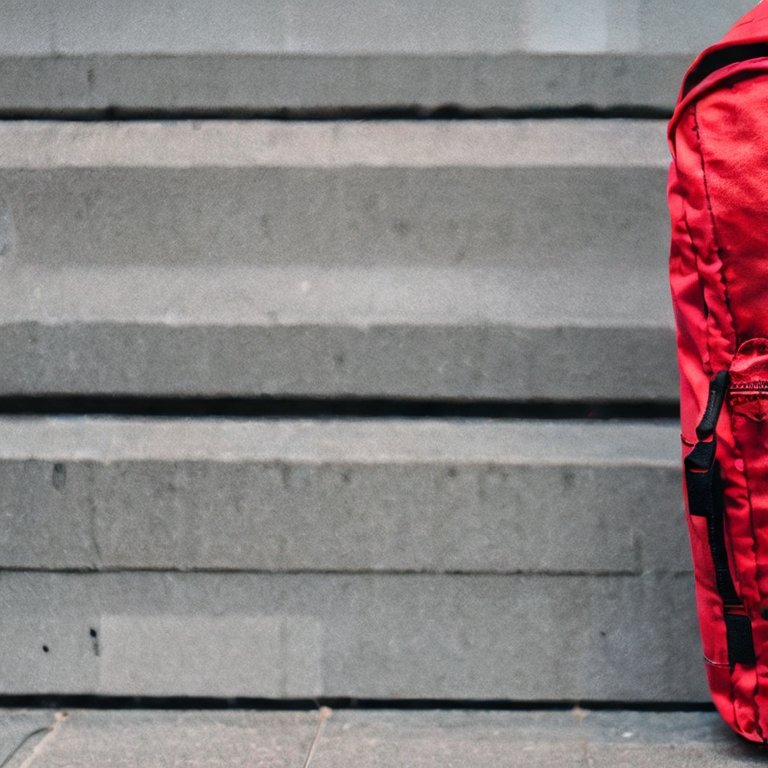}     &
        \includegraphics[width=14mm]{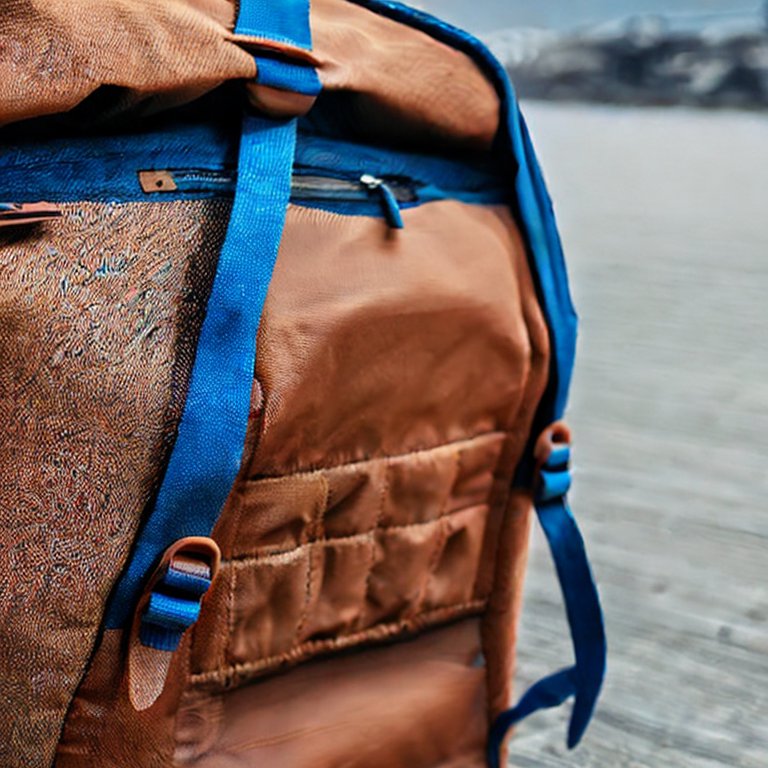}     &
        \includegraphics[width=14mm]{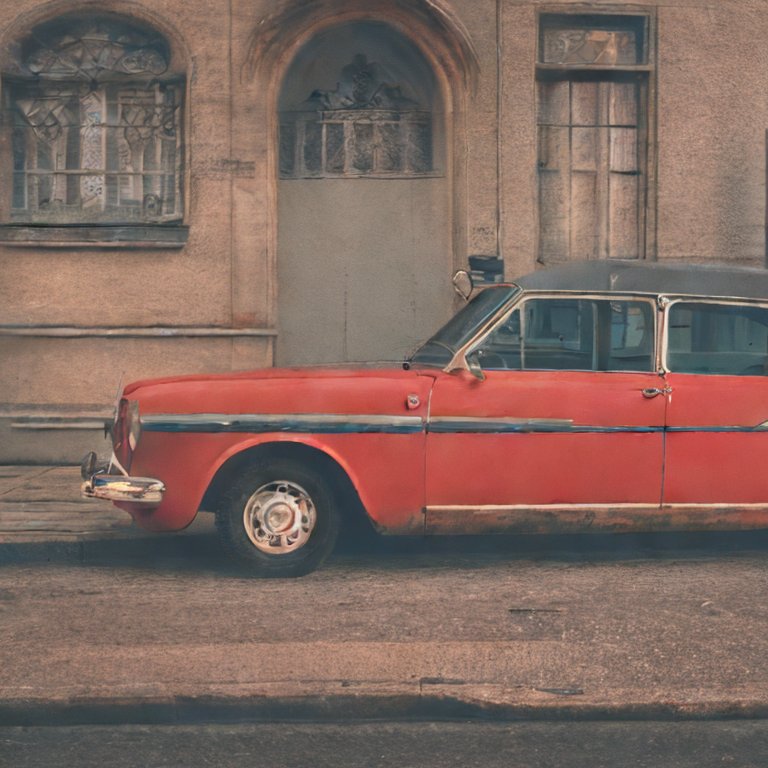}         &
        \includegraphics[width=14mm]{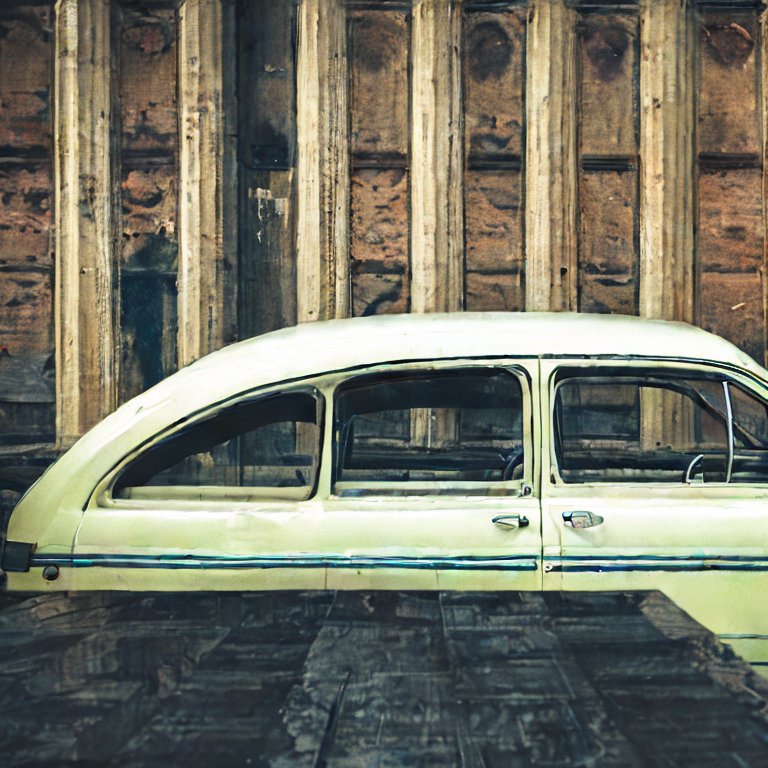}         &
        \includegraphics[width=14mm]{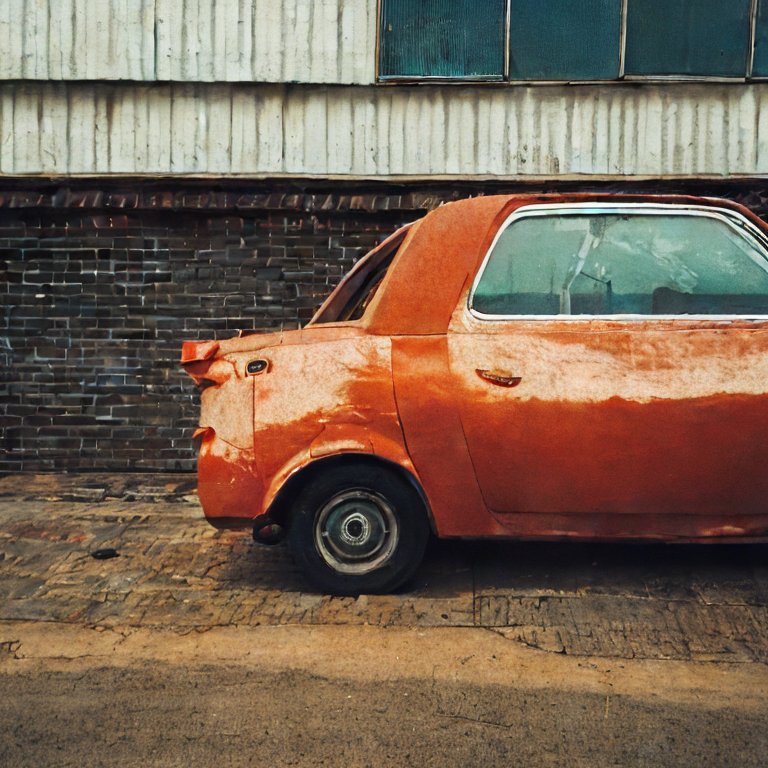}         &
        \includegraphics[width=14mm]{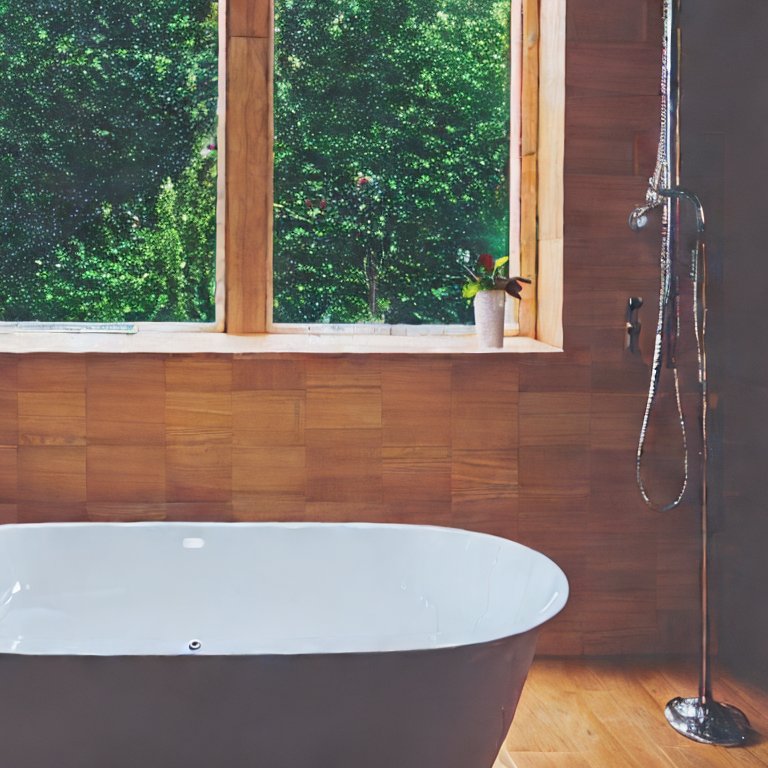}       &
        \includegraphics[width=14mm]{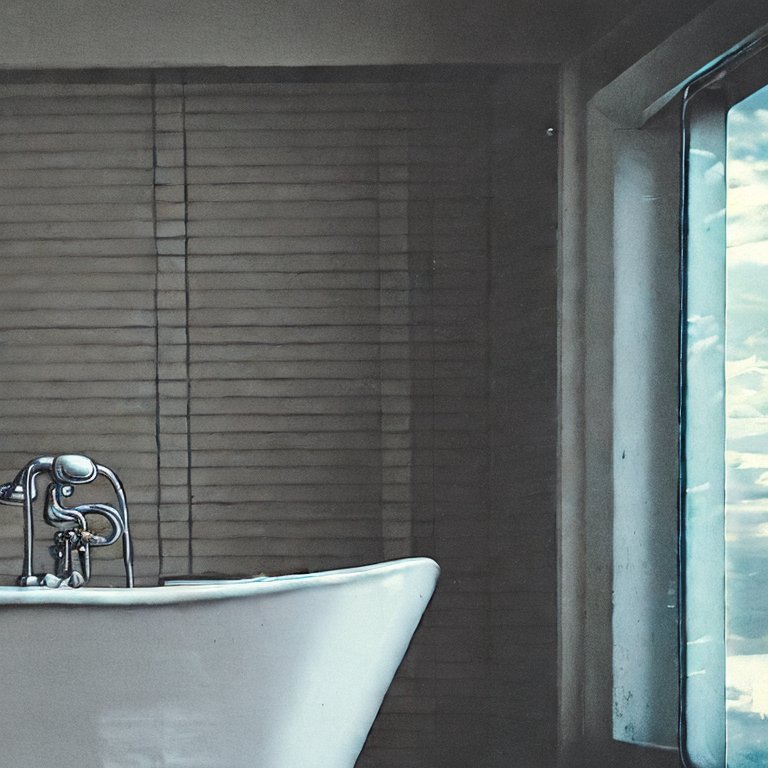}       &
        \includegraphics[width=14mm]{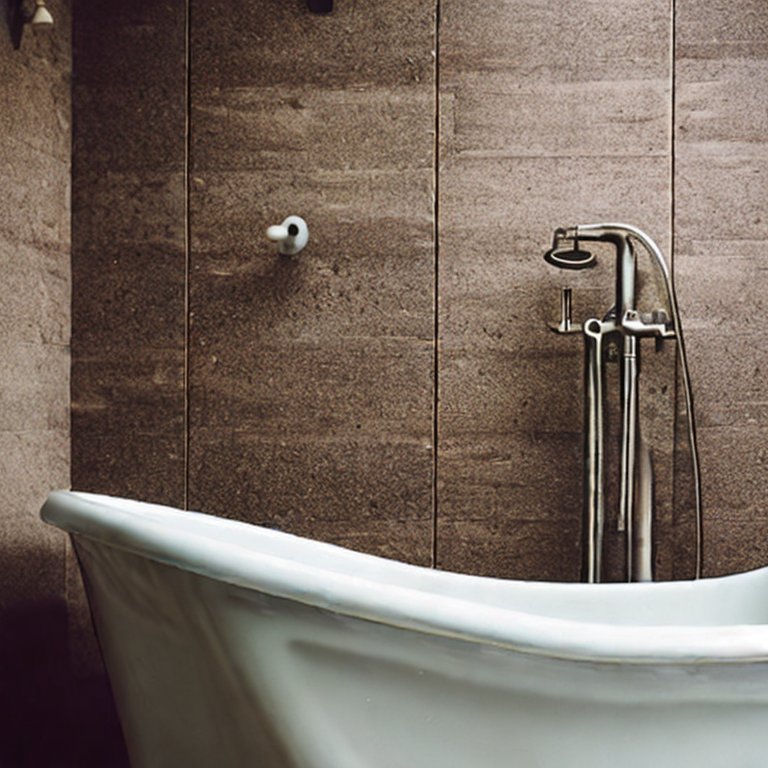}
        \\[-0.3mm]
        \rotatebox[origin=l]{90}{\parbox{13mm}{\centering GLIGEN}}           &
        \includegraphics[width=14mm]{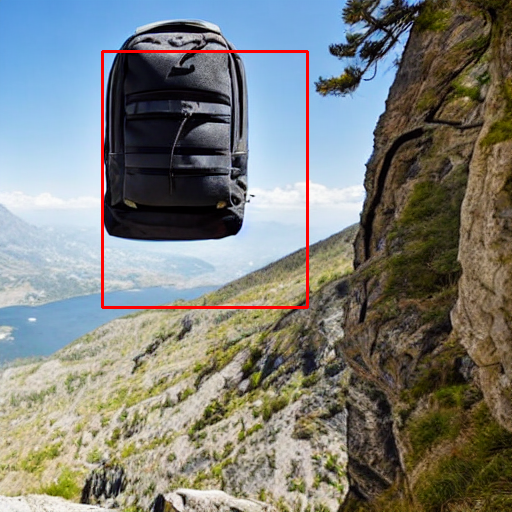} &
        \includegraphics[width=14mm]{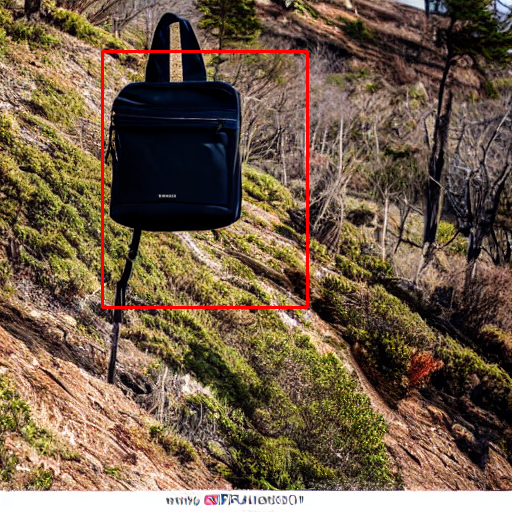} &
        \includegraphics[width=14mm]{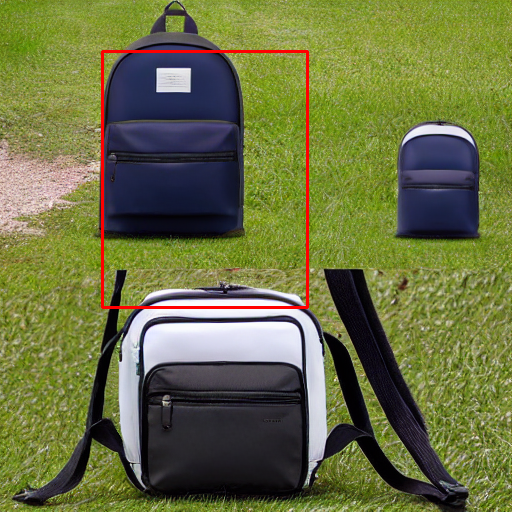} &
        \includegraphics[width=14mm]{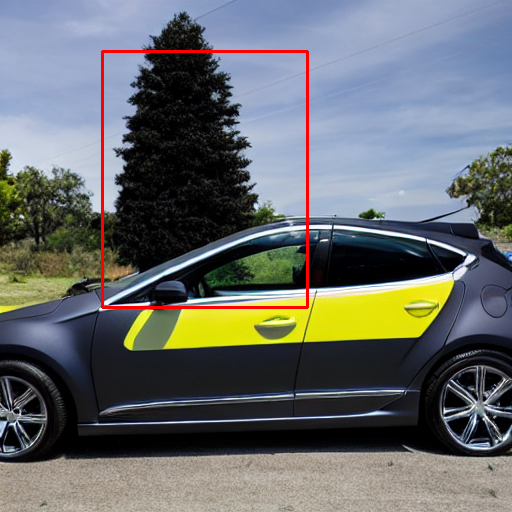}     &
        \includegraphics[width=14mm]{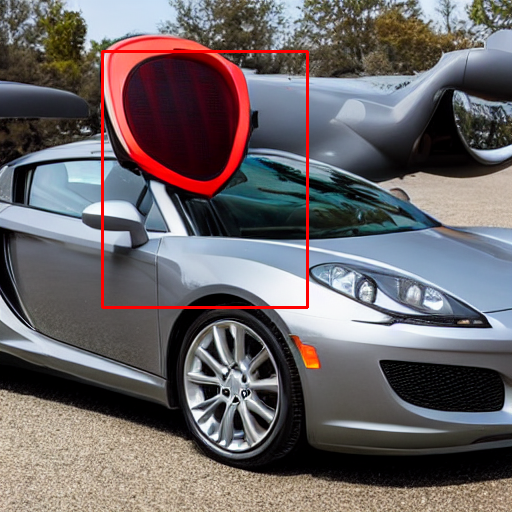}     &
        \includegraphics[width=14mm]{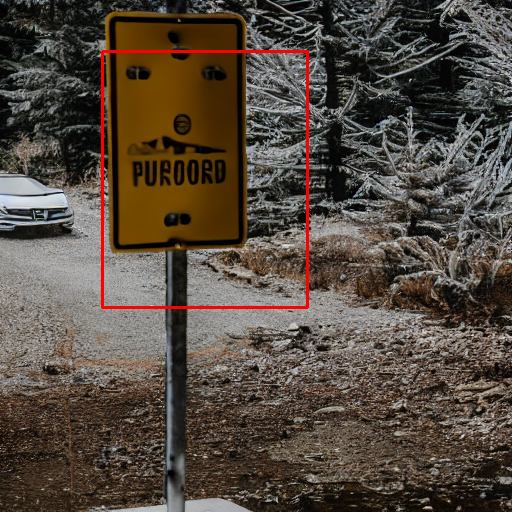}     &
        \includegraphics[width=14mm]{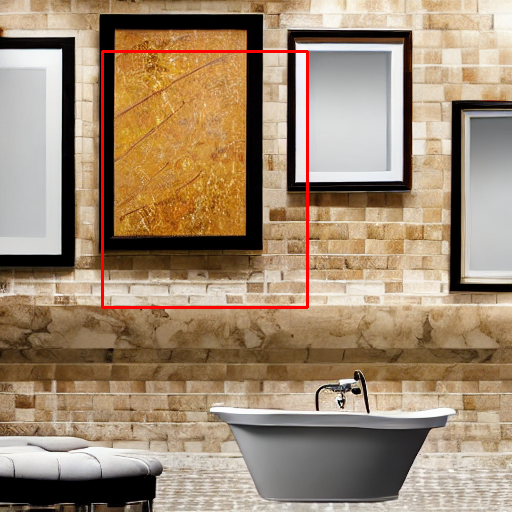}   &
        \includegraphics[width=14mm]{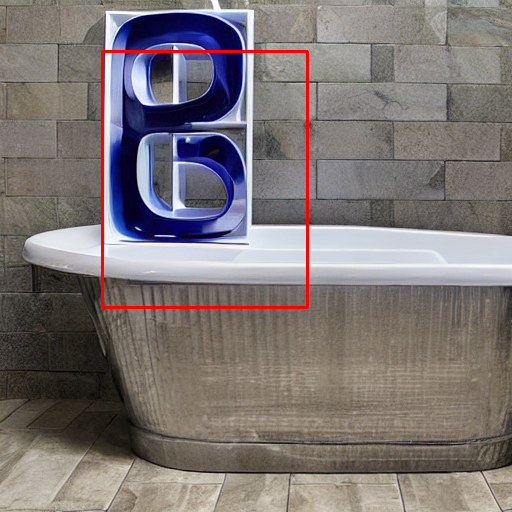}   &
        \includegraphics[width=14mm]{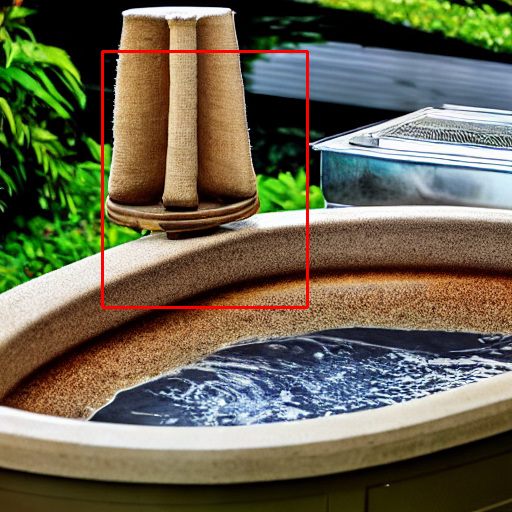}
        \\[-0.3mm]
        \rotatebox[origin=l]{90}{\parbox{13mm}{\centering Layout Guidance}}            &
        \includegraphics[width=14mm]{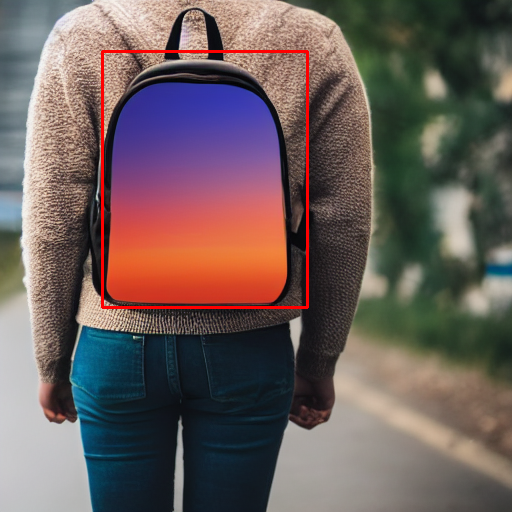} &
        \includegraphics[width=14mm]{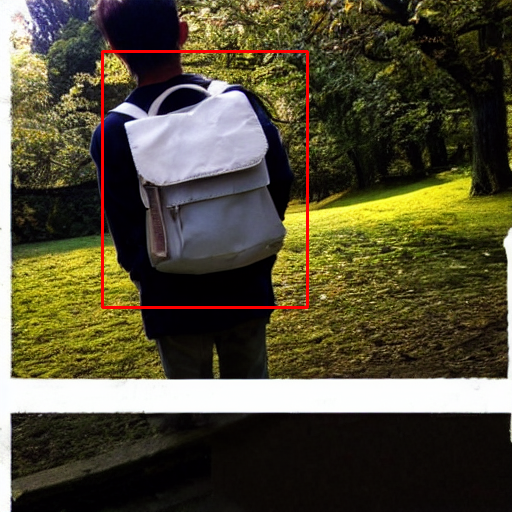} &
        \includegraphics[width=14mm]{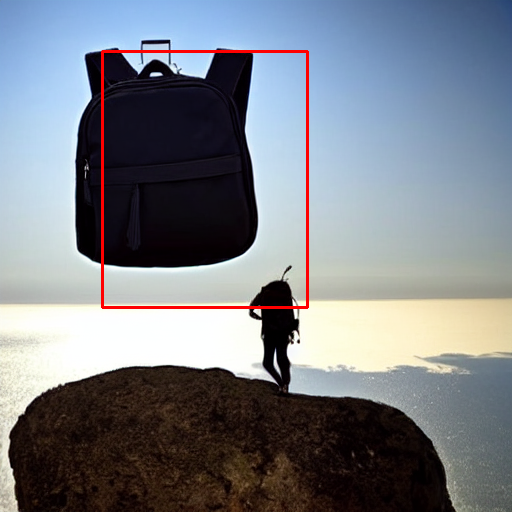} &
        \includegraphics[width=14mm]{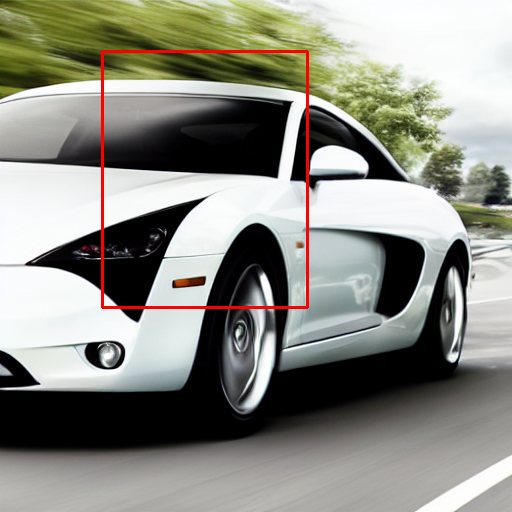} &
        \includegraphics[width=14mm]{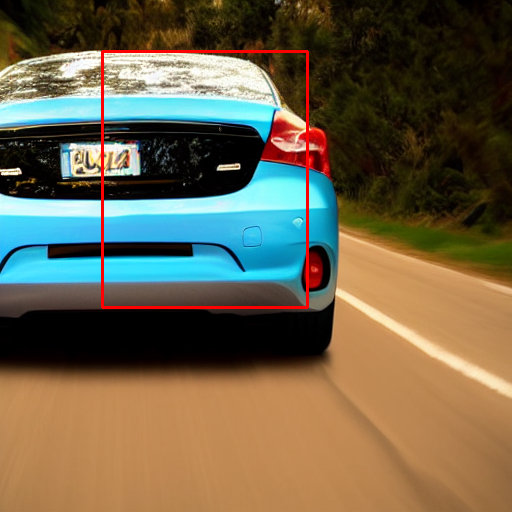}     &
        \includegraphics[width=14mm]{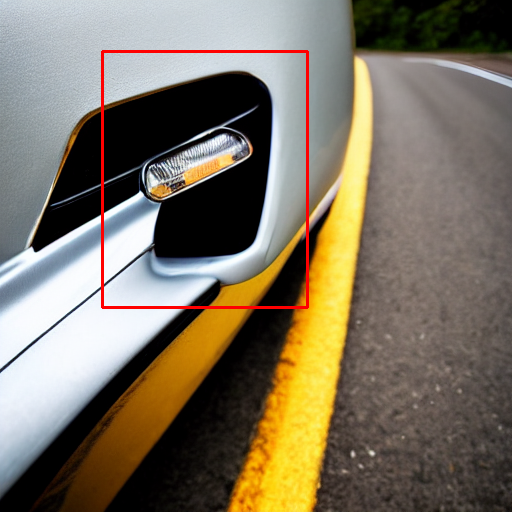}     &
        \includegraphics[width=14mm]{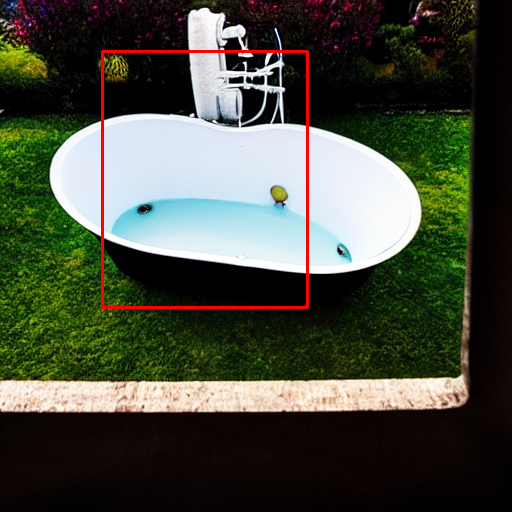}   &
        \includegraphics[width=14mm]{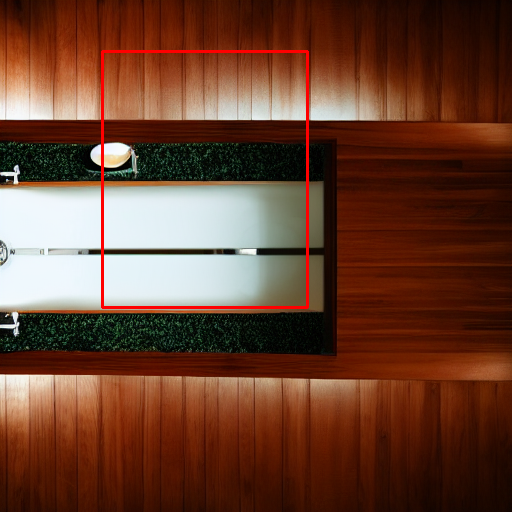}   &
        \includegraphics[width=14mm]{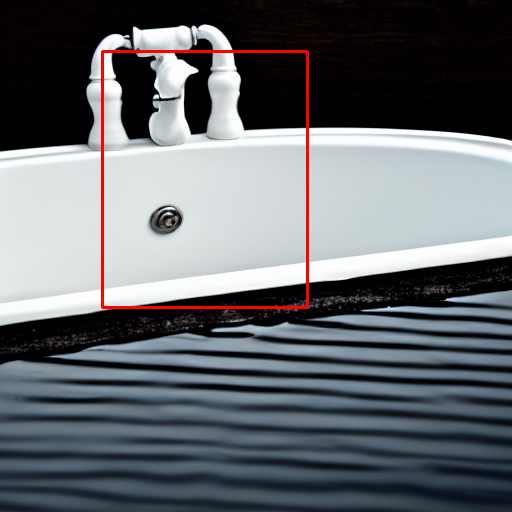}
        \\[-0.3mm]
        \rotatebox[origin=l]{90}{\parbox{13mm}{\centering RPG}}              &
        \includegraphics[width=14mm]{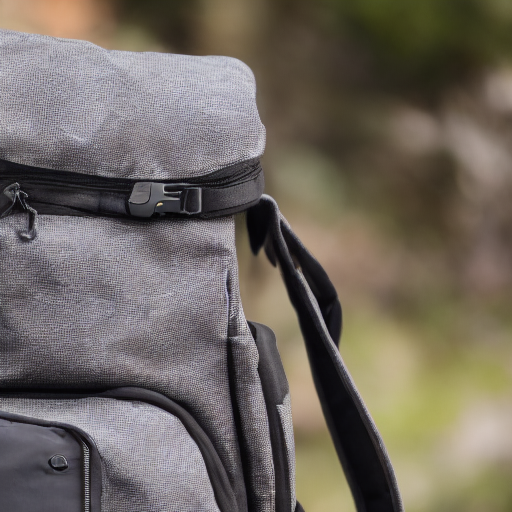}        &
        \includegraphics[width=14mm]{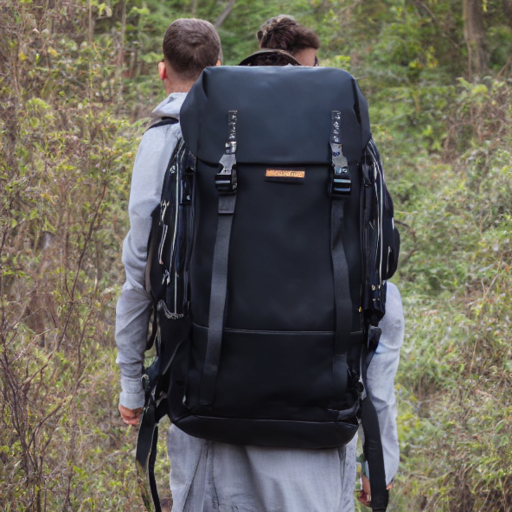}        &
        \includegraphics[width=14mm]{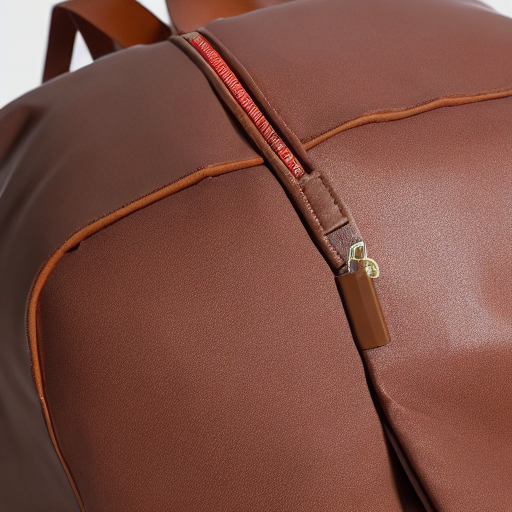}       &
        \includegraphics[width=14mm]{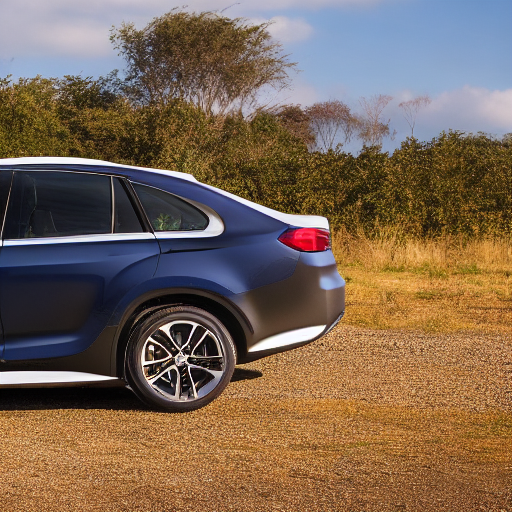}            &
        \includegraphics[width=14mm]{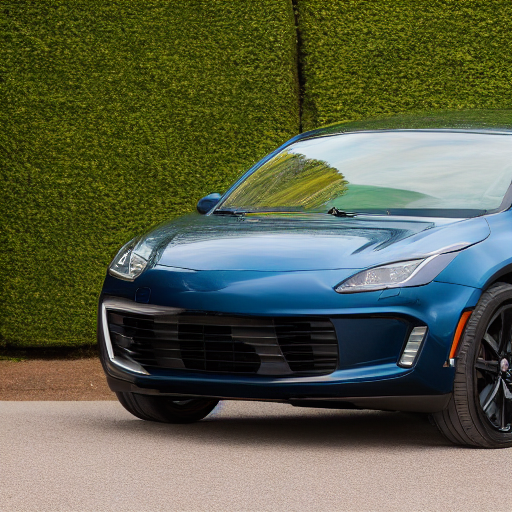}            &
        \includegraphics[width=14mm]{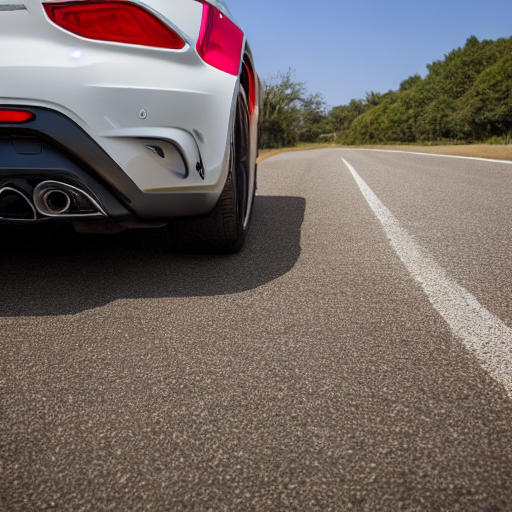}              &
        \includegraphics[width=14mm]{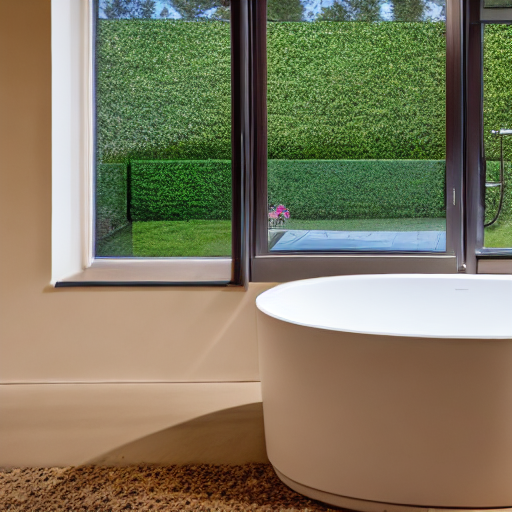}          &
        \includegraphics[width=14mm]{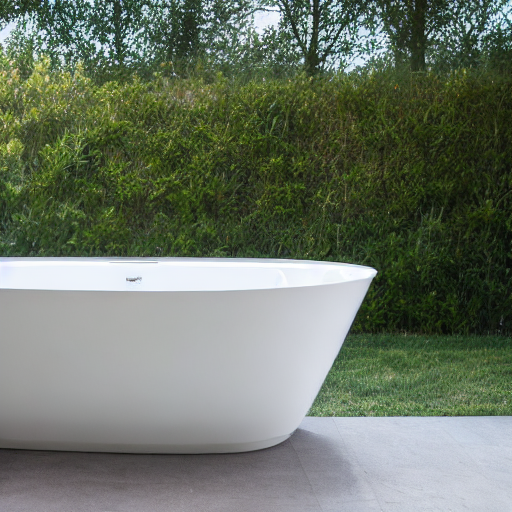}         &
        \includegraphics[width=14mm]{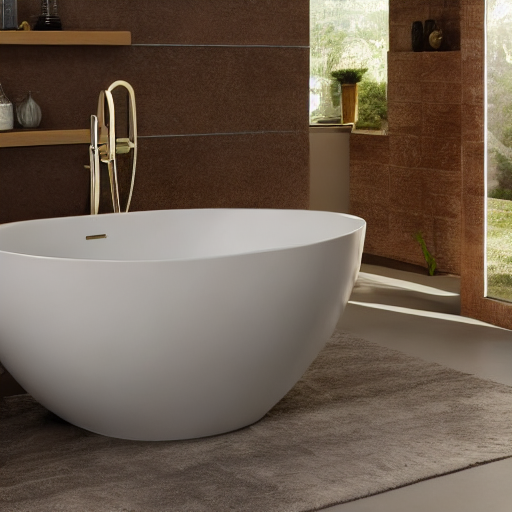} 
        \\[-0.3mm]
        \rotatebox[origin=l]{90}{\parbox{13mm}{\centering \textbf{Ours}}}           &
        \includegraphics[width=14mm]{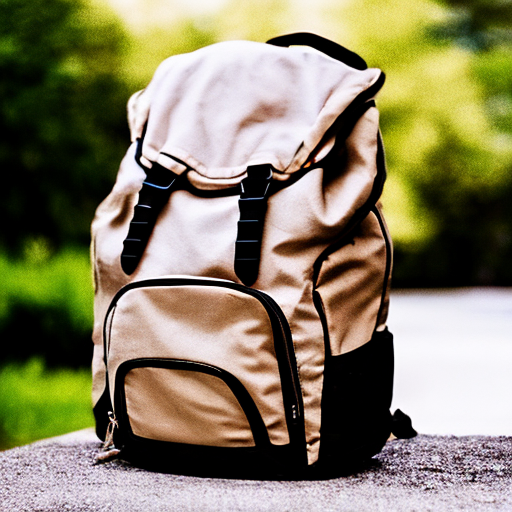} &
        \includegraphics[width=14mm]{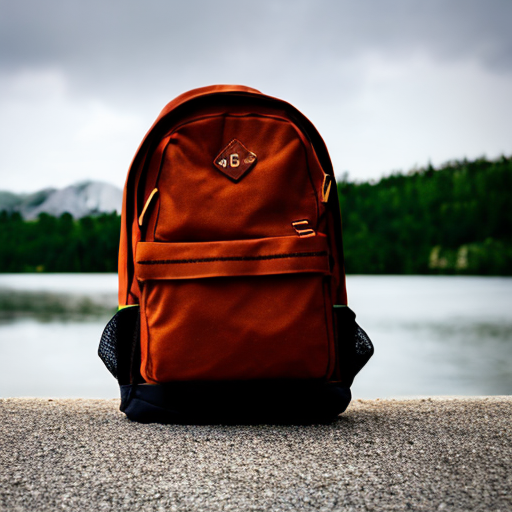} &
        \includegraphics[width=14mm]{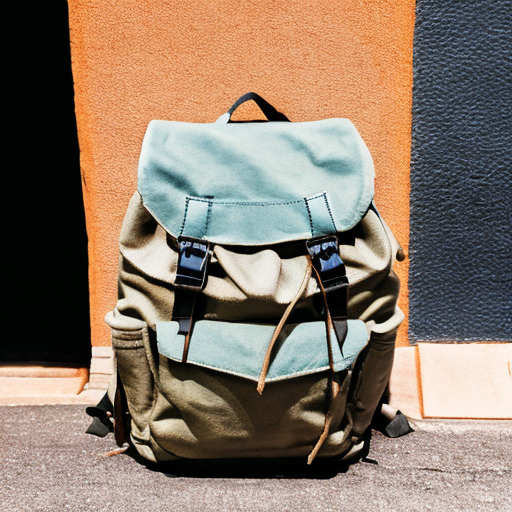} &
        \includegraphics[width=14mm]{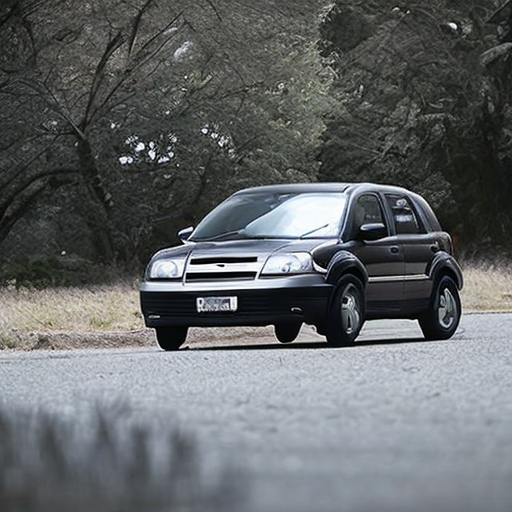}     &
        \includegraphics[width=14mm]{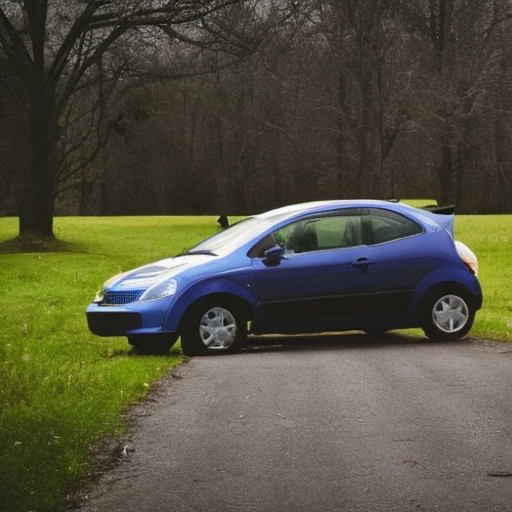}     &
        \includegraphics[width=14mm]{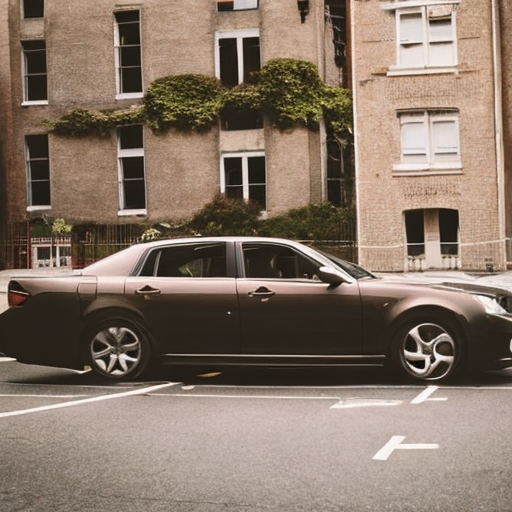}     &
        \includegraphics[width=14mm]{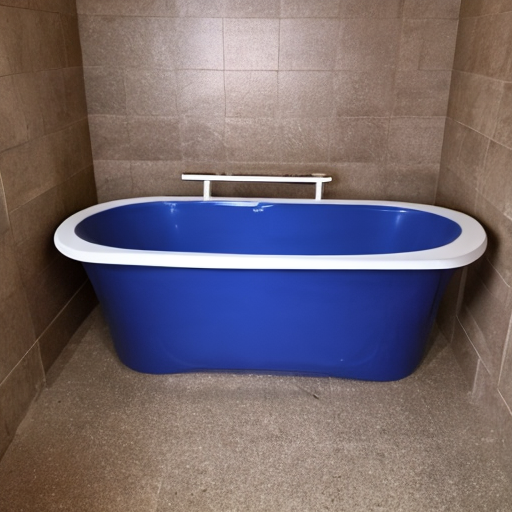}   &
        \includegraphics[width=14mm]{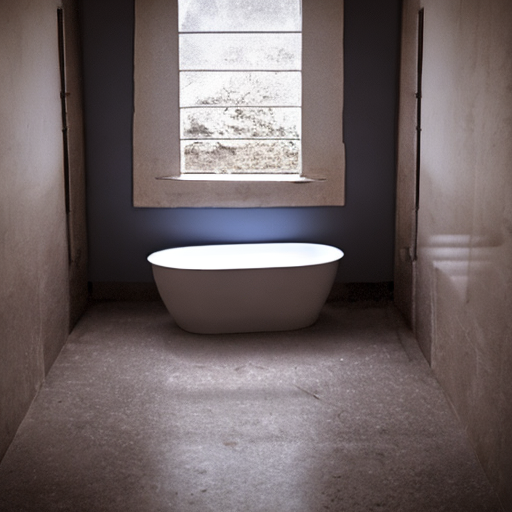}   &
        \includegraphics[width=14mm]{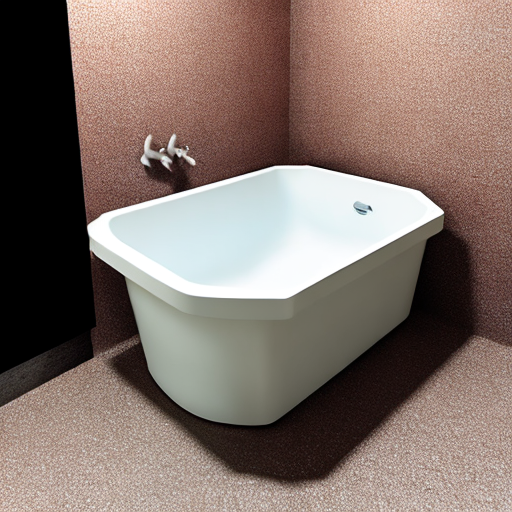}
        \\
    \end{tabular}}
    \captionof{figure}{Visualization comparison between the layout-based methods and our method. For methods that require additional layout information, the bounding box positions are marked on the result images. Our method does not require any additional input.}

    \label{fig:experiment1}
\end{figure*}

\subsection{Evaluation Metrics} 
\label{sec:em}
\textbf{Object Incompleteness Rate.} We have human evaluation and large multimodal models (LMM) evaluation to determine whether one image is complete or not. We calculate the proportion of generated images in which the object is identified as incomplete through human evaluation, referred to as the \textbf{Human-Evaluated Object Incompleteness Rate (HOIR)}. Similar to HOIR, \textbf{LMM-Evaluated Object Incompleteness Rate (LOIR)} leverages LMMs to assess the incompleteness rate on a large scale, primarily using GPT-4~\cite{achiam2023gpt} and LLaVA-v1.5-13B~\cite{liu2023llava} to replace human. The images that are inputted into the LMM are processed with SAM~\cite{kirillov2023segment} to remove background pixels, improving the reliability of the LOIR.\\
\textbf{LMM-Evaluated Completeness Preference Rate (LCPR)} The LCPR metric is derived through a systematic comparison of two images assessed by the LMM: one synthesized using our method and another generated through the baseline approach. We concatenate them and prompt the LMM to determine which appears more complete.\\
\textbf{Time Cost.}  
To assess the computational cost of different methods, we use each model to generate 100 images and calculate the average generation time per image. \\
\textbf{Evaluation with Human Preferences.}
PickScore~\cite{kirstain2023pick}, HPSv2~\cite{wu2023human} and ImageReward~\cite{xu2023imagereward} all serve as evaluation metrics designed to align generated images with human preferences. They assess image quality based on perceptual realism, aesthetic appeal, and alignment with text.\\
\textbf{CLIP-IQA~\cite{wang2023exploring}} quantifies perceptual fidelity through CLIP-encoded feature analysis, evaluating synthesized images across critical dimensions including realism, clarity, and structural coherence.
\begin{figure}[t]
    \centering
    \footnotesize
    \tabcolsep=.1mm
    \begin{tabular}{p{14.5mm}p{17mm}p{17mm}p{17mm}p{17mm}}
    
        \raisebox{2mm}{\parbox{14.5mm}{\centering\quad}}&
        \raisebox{2mm}{\parbox{16mm}{\centering\textit{A bed}}}&
        \raisebox{2mm}{\parbox{16mm}{\centering\textit{A car}}}&
        \raisebox{2mm}{\parbox{16mm}{\centering\textit{A bathtub}}}&
        \raisebox{2mm}{\parbox{16mm}{\centering\textit{A cap}}}\\ 
        
        \raisebox{7mm}{\parbox{14.5mm}{\centering\textbf{Ours w/o cross-attention constraint}}}&
        \includegraphics[width=16mm]{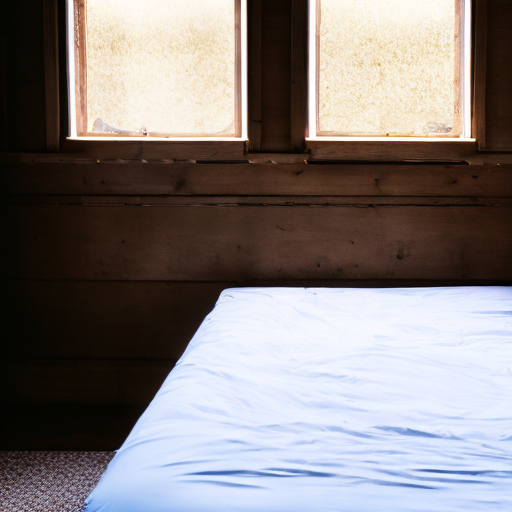}&
        \includegraphics[width=16mm]{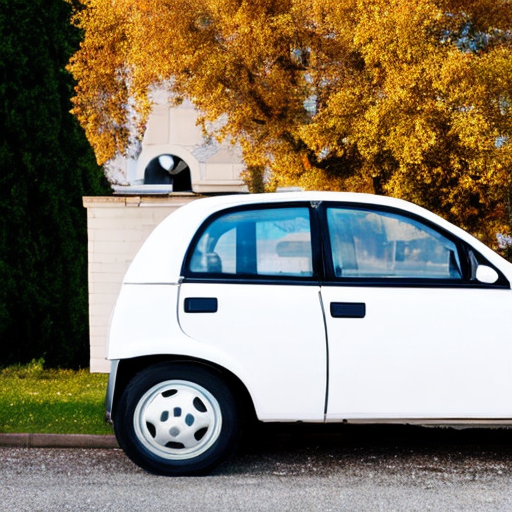}&
        \includegraphics[width=16mm]{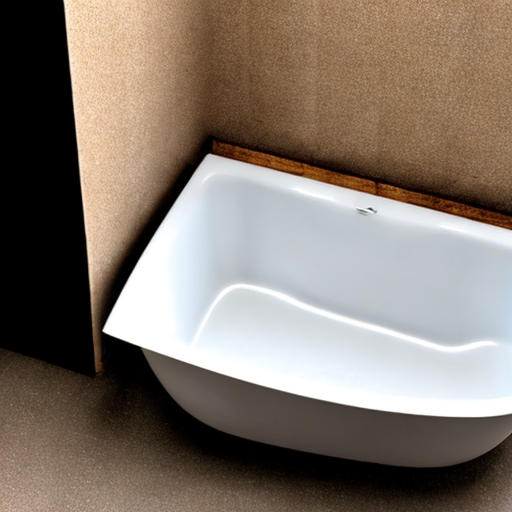}&
        \includegraphics[width=16mm]{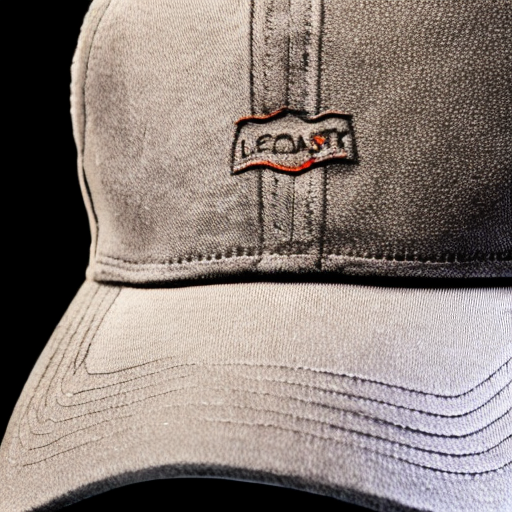}\\

        \raisebox{7mm}{\parbox{14.5mm}{\centering\textbf{Ours w/o self-attention constraint}}}&
        \includegraphics[width=16mm]{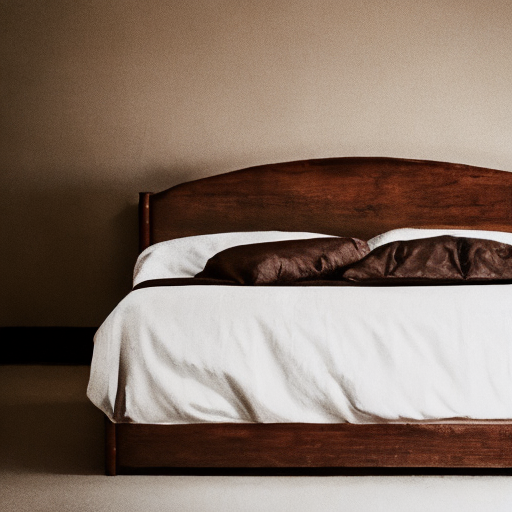}&
        \includegraphics[width=16mm]{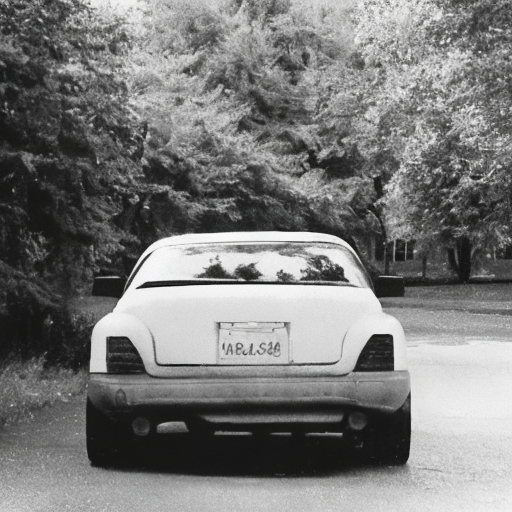}&
        \includegraphics[width=16mm]{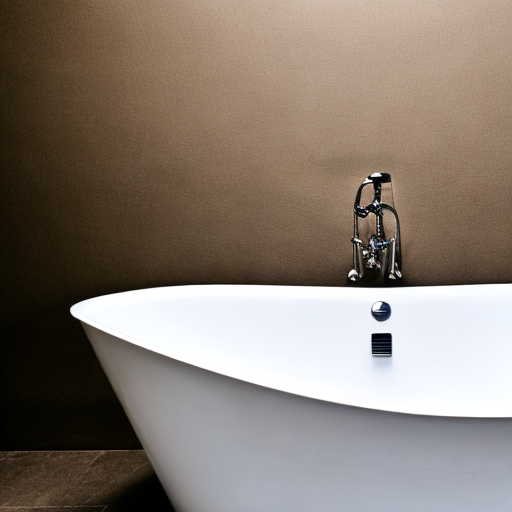}&
        \includegraphics[width=16mm]{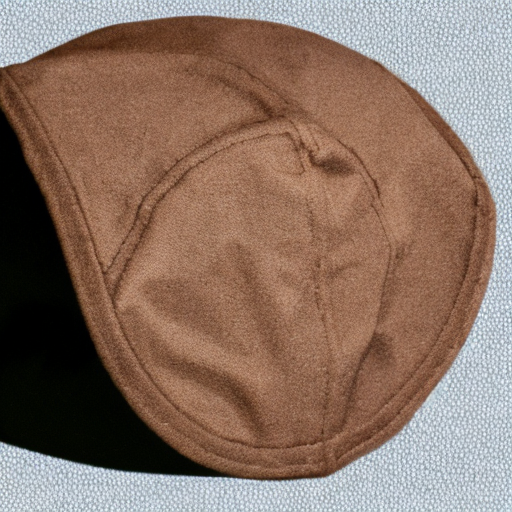}\\

        \raisebox{7mm}{\parbox{14.5mm}{\centering\textbf{Ours}}}&
        \includegraphics[width=16mm]{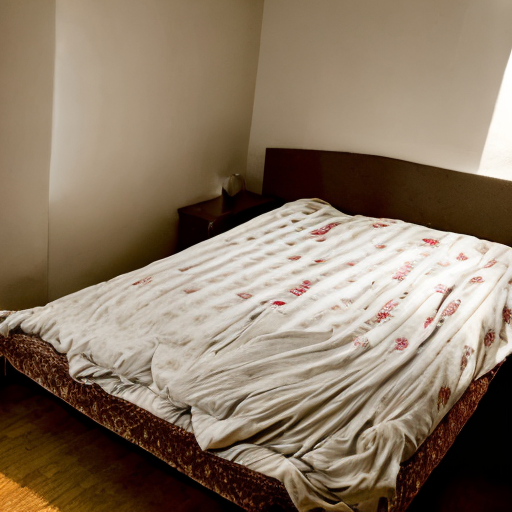}&
        \includegraphics[width=16mm]{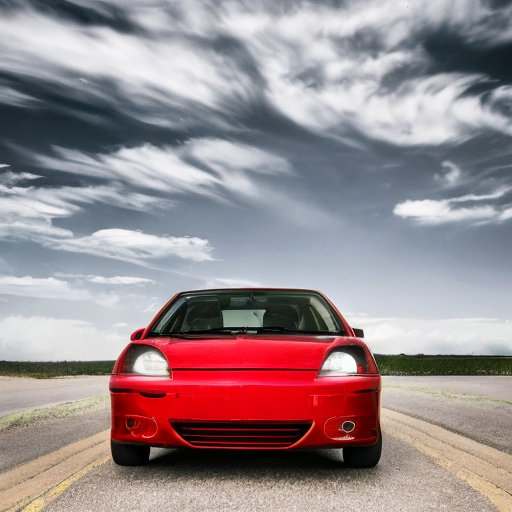}&
        \includegraphics[width=16mm]{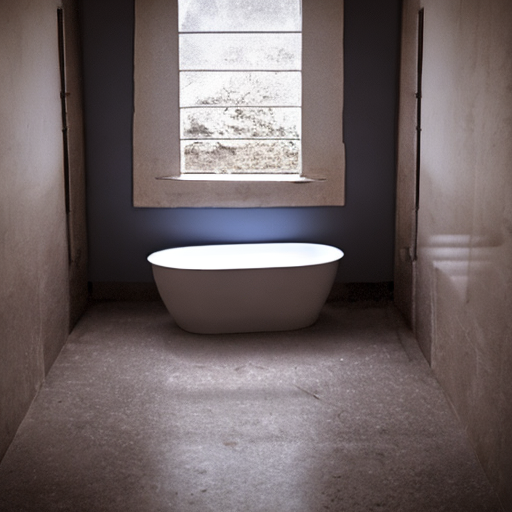}&
        \includegraphics[width=16mm]{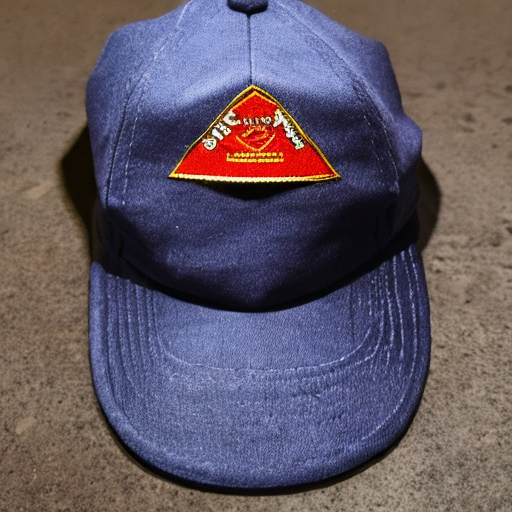}\\
    
    \end{tabular}

    \caption{Visualization of our ablation study.}
    \vspace{-3mm}
    \label{fig:comparison}
\end{figure}
\subsection{Quantitative Comparison}
\label{sec:quantitative}
In this section, we present a quantitative comparison between our proposed method and several layout-based text-to-image generation methods, including GLIGEN~\cite{li2023gligen}, Layout Guidance~\cite{chen2024training}, Boxdiff~\cite{xie2023boxdiff} and four methods~\cite{wu2024self,feng2023layoutgpt,yang2024mastering,qu2023layoutllm} incorporating LLM.\\
\textbf{Object completeness.} As shown in \cref{tab:experiments}, our method achieves best overall performance, with the lowest HOIR and LOIR of 17.3\% and 11.7\%, significantly lower than the baseline Stable Diffusion's 45.57\% and 32.0\%, demonstrating its effectiveness in improving object completeness. Compared to other layout-based methods without LLM, such as GLIGEN~\cite{li2023gligen} and Layout Guidance~\cite{chen2024training}, the best HOIR still reaches as high as 34.2\%, with the corresponding LOIR at 27.9\%. Among LLM-based methods, approaches such as SLG~\cite{wu2024self} and RPG~\cite{yang2024mastering} achieve the best results of only 27.1\% and 23.1\% respectively. As shown in ~\cref{tab:more_bili}, different LMMs consistently assess our method's results as more complete, achieving over 81\% on SDv1.5 and SDv2. Since SDXL produces relatively more complete results, LMMs still favor our method in 71\% and 68\% of cases. As a result, our method still outperforms them in HOIR, LOIR and LCPR.\\
\textbf{Time cost.} Our method exhibits a slight increase compared to Stable Diffusion (5.75s vs. 5.51s), but it is still more efficient than other layout-based methods such as Layout Guidance~\cite{chen2024training} (24.02s) and RPG~\cite{yang2024mastering} (8.54s). This demonstrates that our method significantly enhances image completeness, while the additional computational overhead is negligible.\\
\textbf{Image quality and human preference}, measured by CLIP-IQA~\cite{wang2023exploring}, our method achieves a score of 0.703, which is the closest to Stable Diffusion’s score of 0.714, suggesting that the improvements in object completeness do not severely degrade image quality. In layout generation, conflicts between object placement and the initial layout during denoising may lead to the generation of physically implausible objects, such as the floating backpack in ~\cref{fig:experiment1}, resulting in degraded image quality. \\
When assessing \textbf{image quality with human preference}, our method achieves a significant lead — PickScore~\cite{kirstain2023pick} (23.41 vs. the second-best 23.07), HPSv2~\cite{wu2023human} (25.66 vs. 25.17), and ImageReward~\cite{xu2023imagereward} (0.327 vs. 0.253). These results suggest that human preference favors complete object representations, demonstrating our method's effectiveness.

\begin{table}[t]
    \footnotesize
    \centering
    \resizebox{\linewidth}{!}{
    \begin{tabular}{l|ccc}
        \toprule
        \textbf{Model} & {\textbf{HOIR $\downarrow$} }  & {\textbf{LCPR (GPT4) $\uparrow$} } & {\textbf{LCPR (LLaVA) $\uparrow$} }\\
        \midrule
        SD1.5           & 43.7 \%  & - & -  \\
        \rowcolor{gray!20}
        SD1.5 + Ours    & 18.7 \%  & 81 \% & 88 \%      \\
        
        \midrule
        SD2            & 45.5 \%   & -  & - \\
        \rowcolor{gray!20}
        SD2 + Ours     & 17.3 \%   & 85 \% & 81 \%   \\

        \midrule
        SDXL            & 18.3 \%    & -  & -  \\
        \rowcolor{gray!20}
        SDXL + Ours     & 7.1 \%   & 71 \%  & 68 \%   \\
        \bottomrule
    \end{tabular}}
    \vspace{-1mm}
    \caption{Comparison of incompleteness rates across different models with our method applied. In this experiment, we tested 1,200 prompts. 360 prompts contain descriptions of 2 to 5 objects, while the remaining 840 include only a single object. The higher the LCPR, the more broadly and effectively our method enhances image completeness.}
    \label{tab:more_bili}

\end{table}

\subsection{Visualization Comparison}
We present a visualization comparison between our method with the layout-based methods in ~\cref{fig:experiment1}. \\
\textbf{Object completeness.} Our results clearly show complete details of the object as described in the prompt, such as details like the straps of a backpack or the rear end of a car. As shown in~\cref{fig:experiment1}, for the compared layout-to-image methods, objects in the generated images often exceed the given box boundaries (such as the car in the fifth column of GLIGEN, in the fourth column of Layout Guidance, etc.). The analysis in ~\cref{{sec:quantitative}} and the complete objects shown in ~\cref{fig:experiment1} demonstrate the feasibility of our method. \\
\textbf{Image realism and quality.}
Our method yields peak performance in image realism and quality due to the preservation of fine, complete details and the natural placement of objects in appropriate positions.
As shown in~\cref{fig:experiment1}, for the comparison methods, generated objects often appear unrealistic despite being complete, such as the floating backpack in the second row. This aligns with CLIP-IQA results and human preference. Other issues include distorted objects, anomalous textures (e.g., the third row's first backpack), and inconsistent regional styles (e.g., the third row's second backpack with mismatched upper and lower halves).

In summary, while layout-guidance can address the problem of object incompleteness from another perspective, some of its results come at the cost of image realism and quality. Our method seldom produces results where the objects are in extreme discord with the background. 

\begin{table}[t]
    \footnotesize
    \centering
    \resizebox{\linewidth}{!}{
    \begin{tabular}{lccc}
        \toprule
        \textbf{Method} & {\textbf{HOIR $\downarrow$} } & {\textbf{LOIR $\downarrow$} } & \textbf{CLIP-IQA $\uparrow$}    \\
        \midrule
        Ours w/o  Cross Constraint           & 38.7 \%     &  24.5\%           & 0.643 \\
        Ours w/o  Self Constraint            & 34.7 \%     &  26.9\%         & 0.681\\
        \midrule
        \textbf{Ours}            &  \textbf{17.3 \%}  & \textbf{11.7\%}  & \textbf{0.700} \\
        \bottomrule
    \end{tabular}}

    \caption{Ablation study on whether to use the cross-attention constraint and self-attention constraint of our method.
    }
    \label{tab:experiments_ablation}

\end{table}

\begin{figure}[t]
    \centering
    \includegraphics[width=.40\textwidth]{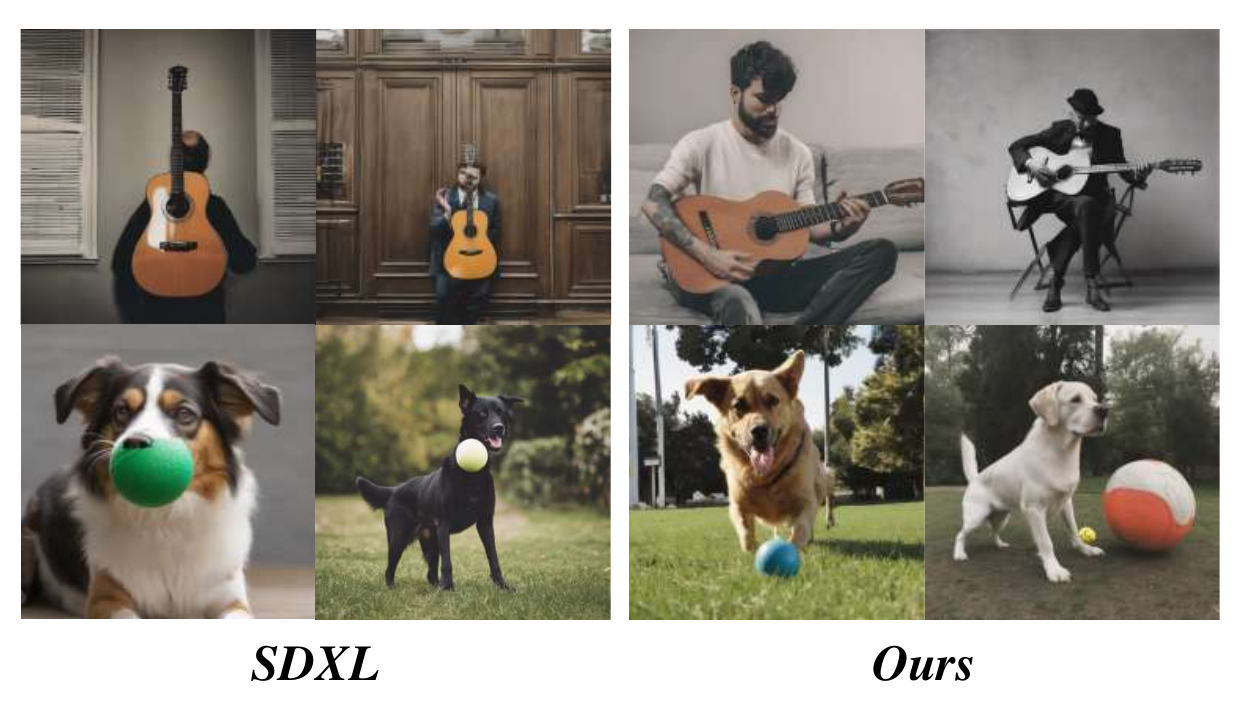}
    
    \caption{Visualization of SDXL generation results. SDXL sometimes enforces object centering.}
    \label{fig:sdxlcomparison}

\end{figure}

\subsection{Comparison Analysis of SDXL}\label{sec:sdxl} 
SDXL addresses incompleteness issue by incorporating the coordinates of the cropped operation during training. In contrast, our method can be summarized in the following three aspects. 
\textbf{1) Orthogonal approach.} Our approach is orthogonal to the training method proposed by SDXL and can be easily integrated into SDXL. In fact, training or fine-tuning is prohibitively expensive, whereas our approach serves as an effective remedy for widely deployed pretrained models. 
\textbf{2) Further progress.} 
As shown in ~\cref{tab:more_bili}, SDXL exhibits the HOIR of 18.0\%. However, after incorporating our method, the HOIR is reduced to 7.1 \%, demonstrating the effectiveness of our approach in further addressing the incompleteness issue. 
\textbf{3) Avoiding minor shortcomings in SDXL.} 
The method proposed by SDXL addresses the incompleteness issue by incorporating the top-left crop coordinates during training. This seems to cause the model to forcibly center objects. For a single object, it is guided to an appropriate position, but for multiple objects, it rigidly adheres to the placement of a single object’s coordinates. As shown in the left two columns of ~\cref{fig:sdxlcomparison}, the guitar and player, as well as the dog and ball, become entangled. In contrast, our method pushes objects away from the boundaries, ensuring consistent effectiveness for both single and multiple objects. 

\subsection{Ablation Study}
\label{sec:Ablation}
\textbf{Impact of the cross-attention constraint.}
We conduct experiments under the same parameter settings without the cross-attention constraint. As shown in the first row of~\cref{tab:experiments_ablation}, with the absence of $\mathcal L^{\texttt{cross}}$, the rate of incomplete results increased by 21.4\%. 
The results in the first row of~\cref{fig:comparison} show that without cross-attention constraint, the objects still appear incomplete in the generated images. Such as the car in results lost its front part. 
It proves that relying solely on self-attention guidance is inadequate for accurately associating the object’s generated position with its semantic expressions. This conclusion is consistent with the greater decline in CLIP-IQA metrics in~\cref{tab:experiments_ablation}. \\
\textbf{Impact of the self-attention constraint.}
When using only cross-attention guidance, as shown in ~\cref{tab:experiments_ablation}, the incompleteness rate increases by 17.4\%. In the second row of ~\cref{fig:comparison}, objects like the bathtub and hat persist at image boundaries, indicating self-attention's critical role in position refinement. 

\begin{figure}[t]
    \centering
    \footnotesize
    \tabcolsep=.1mm
    \begin{tabular}{p{14.5mm}p{17mm}p{17mm}p{17mm}p{17mm}}
    
        \raisebox{1mm}{\parbox{14.5mm}{\centering\quad}}&
        \raisebox{1mm}{\parbox{16mm}{\centering\textbf{T=940}}}&
        \raisebox{1mm}{\parbox{16mm}{\centering\textbf{T=900}}}&
        \raisebox{1mm}{\parbox{16mm}{\centering\textbf{T=1}}}&
        \raisebox{1mm}{\parbox{16mm}{\centering\textbf{Image}}}\\
        
        \raisebox{7mm}{\parbox{14.5mm}{\centering\textbf{Stable Diffusion}}}&
        \includegraphics[width=16mm]{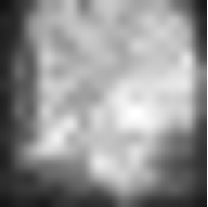}&
        \includegraphics[width=16mm]{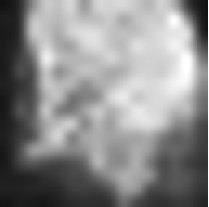}&
        \includegraphics[width=16mm]{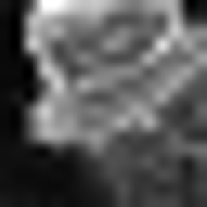}&
        \includegraphics[width=16mm]{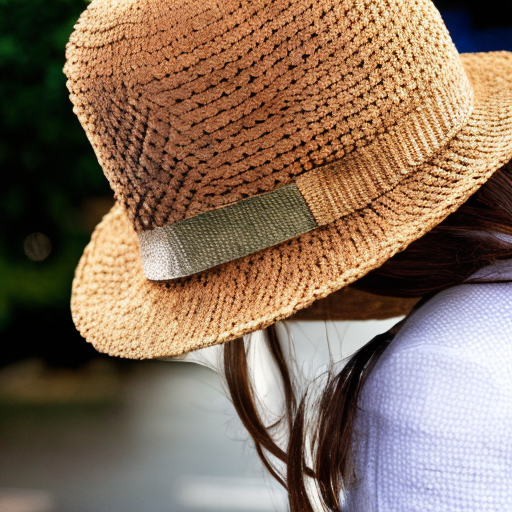}\\
    
        \raisebox{7mm}{\parbox{14.5mm}{\centering\textbf{w/o cross-attention constraint}}}&
        \includegraphics[width=16mm]{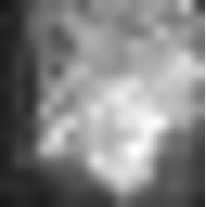}&
        \includegraphics[width=16mm]{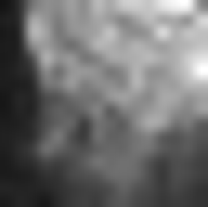}&
        \includegraphics[width=16mm]{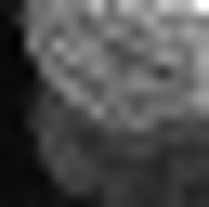}&
        \includegraphics[width=16mm]{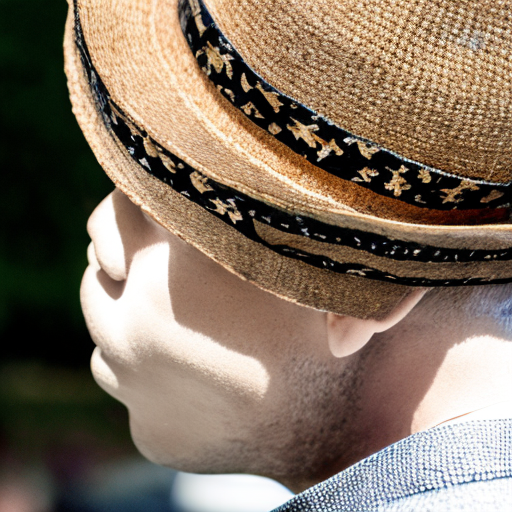}\\

        \raisebox{7mm}{\parbox{14.5mm}{\centering\textbf{w/o self-attention constraint}}}&
        \includegraphics[width=16mm]{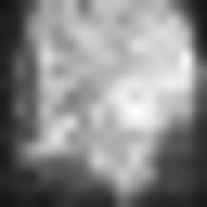}&
        \includegraphics[width=16mm]{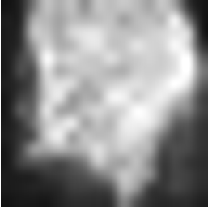}&
        \includegraphics[width=16mm]{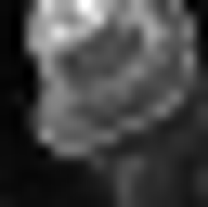}&
        \includegraphics[width=16mm]{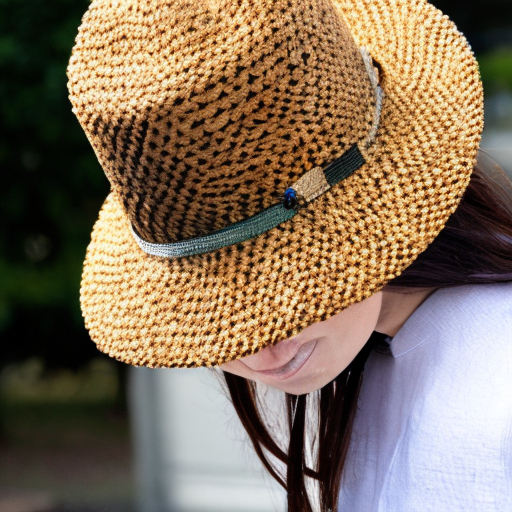}\\

        \raisebox{7mm}{\parbox{14.5mm}{\centering\textbf{Ours}}}&
        \includegraphics[width=16mm]{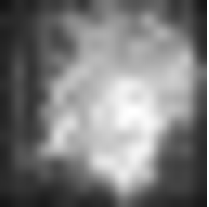}&
        \includegraphics[width=16mm]{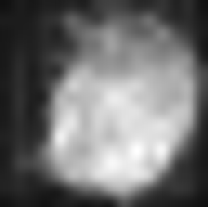}&
        \includegraphics[width=16mm]{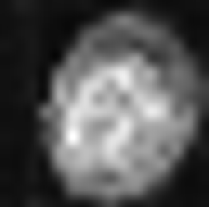}&
        \includegraphics[width=16mm]{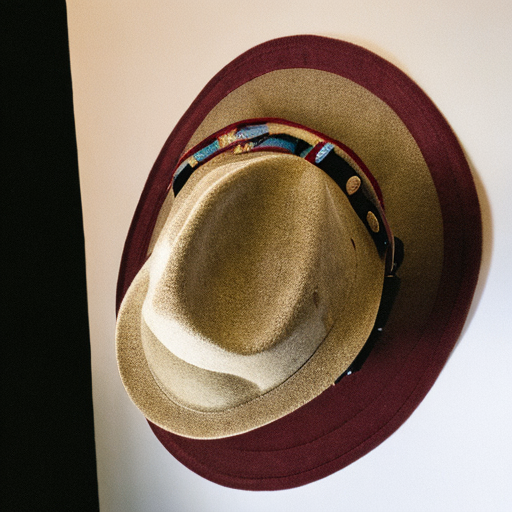}\\
    
    \end{tabular}
    \vspace{-2mm}
    \caption{Ablation study on the variation of attention maps across time steps.}
    \label{fig:attention}
\end{figure}

These findings validate that both cross-attention and self-attention constraints complement each other, with self-attention improving the localization and structure of objects, while cross-attention focuses on enhancing object completeness and detail. As evidenced by the attention map changes in ~\cref{fig:attention}, the combined guidance of both can efficiently correct the tendency for incomplete generation.    

\section{Conclusion}
\noindent In this study, we identified a significant issue in diffusion models: the incomplete display of objects in generated images. Through our in-depth analysis, we found that using \textit{RandomCrop} during training is a major contributor to this issue. To address this, we proposed a training-free method that can be easily applied to pre-trained models like Stable Diffusion. Our method enhances object completeness and improves the overall quality of generated images with minimal computational cost. Extensive experiments demonstrated the effectiveness of our method, showing significant improvements in object completeness and image realism. This work not only provides a practical solution for improving the performance of diffusion models but also contributes to the development of more reliable and accurate text-to-image generation techniques for real-world applications.
\section*{Acknowledgements}
This work was supported in part by the National Natural Science Foundation of China under grant 62471344, in part by the Key Research and Development Program of Yunnan Province under Grant No. 202403AA080002, and in part by the National Key Research and Development Program of China under grant 2023YFC2705700.
{
    \small
    \bibliographystyle{ieeenat_fullname}
    \bibliography{main}
}

\end{document}